\documentclass{article} 
\usepackage{iclr2021_conference,times}

\usepackage{amsmath,amsfonts,bm}

\def\eqref#1{equation~\ref{#1}}

\def\1{\bm{1}}

\DeclareMathAlphabet{\mathsfit}{\encodingdefault}{\sfdefault}{m}{sl}
\SetMathAlphabet{\mathsfit}{bold}{\encodingdefault}{\sfdefault}{bx}{n}

\usepackage{hyperref}
\usepackage{url}

\usepackage{graphicx}
\usepackage{tabularx}
\usepackage{wrapfig,booktabs}
\usepackage{amsmath,amssymb} 
\usepackage{array,multirow}

\newcommand{\STAB}[1]{\begin{tabular}{@{}c@{}}#1\end{tabular}}

\title{Group Equivariant Generative Adversarial Networks}

\author{Neel Dey\thanks{Work started and partially done during an internship at Merck \& Co., Inc.}\\
New York University\\
{\small \texttt{neel.dey@nyu.edu}} \\
\And
Antong Chen \& Soheil Ghafurian\thanks{Work done while employed at Merck \& Co., Inc.} \\
Data Science \& Scientific Informatics, Merck \& Co., Inc. \\
{\small \texttt{antong.chen@merck.com}, \  \texttt{soheilghafurian@gmail.com}}
}

\iclrfinalcopy 
\begin{document}

\maketitle

\begin{abstract}
Recent improvements in generative adversarial visual synthesis incorporate real and fake image transformation in a self-supervised setting, leading to increased stability and perceptual fidelity. However, these approaches typically involve image augmentations via additional regularizers in the GAN objective and thus spend valuable network capacity towards approximating transformation equivariance instead of their desired task. In this work, we explicitly incorporate inductive symmetry priors into the network architectures via group-equivariant convolutional networks. Group-convolutions have higher expressive power with fewer samples and lead to better gradient feedback between generator and discriminator.
We show that group-equivariance integrates seamlessly with recent techniques for GAN training across regularizers, architectures, and loss functions. We demonstrate the utility of our methods for conditional synthesis by improving generation in the limited data regime across symmetric imaging datasets and even find benefits for natural images with preferred orientation.

\end{abstract}

\section{Introduction}

Generative visual modeling is an area of active research, time and again finding diverse and creative applications. A prevailing approach is the generative adversarial network (GAN), wherein density estimation is implicitly approximated by a min-max game between two neural networks \citep{goodfellow2014generative}. Recent GANs are capable of high-quality natural image synthesis and scale dramatically with increases in data and compute \citep{brock2018large}. However, GANs are prone to instability due to the difficulty of achieving a local equilibrium between the two networks. Frequent failures include one or both networks diverging or the generator only capturing a few modes of the empirical distribution. Several proposed remedies include modifying training objectives \citep{arjovsky2017wasserstein,jolicoeur2018relativistic}, hierarchical methods \citep{karras2017progressive}, instance selection \citep{sinha2019small,sinha2020top}, latent optimization \citep{wu2019logan}, and strongly regularizing one or both networks \citep{gulrajani2017improved,miyato2018spectral,dieng2019prescribed}, among others. In practice, one or all of the above techniques are ultimately adapted to specific use cases.

Further, limits on data quantity empirically exacerbate training stability issues more often due to discriminator overfitting. Recent work on GANs for small sample sizes can be roughly divided into transfer learning approaches \citep{wang2018transferring,noguchi2019image,mo2020freeze,zhao2020leveraging} or methods which transform/augment the available training data and provide the discriminator with auxiliary tasks. For example, \citet{chen2019self} propose a multi-task discriminator which additionally predicts the degree by which an input image has been rotated, whereas \citet{Zhang2020Consistency,zhao2020improved} incorporate consistency regularization where the discriminator is penalized towards similar activations for transformed/augmented real and fake images. However, with consistency regularization and augmentation, network capacity is spent learning equivariance to transformation as opposed to the desired task and equivariance is not guaranteed.

\begin{figure}[t]
    \centering
    \includegraphics[width=0.85\textwidth]{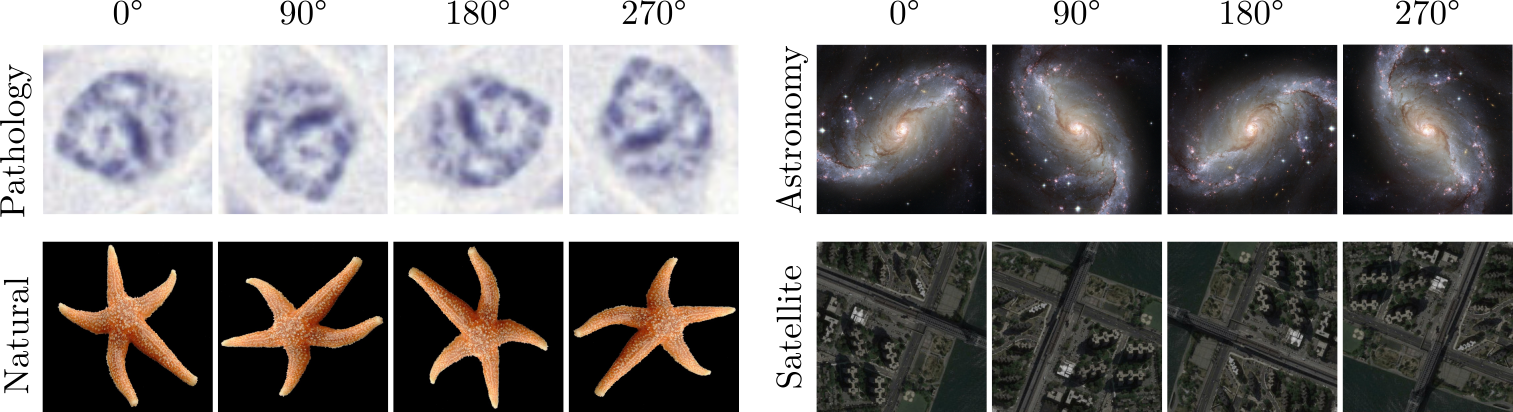}
    \caption{Several image modalities have no preferred orientation for tasks such as classification. We improve their generative modeling by utilizing image symmetries within a GAN framework. 
    }
    \label{fig:intro}
\end{figure}

In this work, we consider the problem of training \textit{tabula rasa} on limited data which possess global and even local symmetries. We begin by noting that GANs ubiquitously use convolutional layers which exploit the approximate translation invariance and equivariance of image labels and distributions, respectively. Equivariance to geometric transformations is key to understanding image representations
\citep{bietti2019group}.
Unfortunately, other symmetries (e.g., rotations and reflections) inherent to modalities such as astronomy and medical imaging where galaxies and cells can be in arbitrary orientations are not accounted for by standard convolutional layers. To this end, \citet{cohen2016group} proposed a group-theoretic generalization of convolutional layers (group-convolutions) which in addition to translation, exploit other inherent symmetries and increase the expressive capacity of a network thereby increasing its sample efficiency significantly in detection \citep{winkels2019pulmonary}, classification \citep{veeling2018rotation}, and segmentation \citep{chidester2019enhanced}. Importantly, equivariant networks outperform standard CNNs trained with augmentations from the corresponding group \citep[Table 1]{veeling2018rotation}, \citep[Fig. 7]{lafarge2020rototranslation}.
See \citet{cohen2019general,esteves2020theoretical} for a formal treatment of equivariant CNNs. 

Equivariant features may also be constructed via scattering networks consisting of non-trainable Wavelet filters, enabling equivariance to diverse symmetries \citep{mallat2012scattering, bruna2013scattering, Sifre_2013_CVPR}. Generative scattering networks include \cite{angles2018generative} where a standard convolutional decoder is optimized to reconstruct images from an embedding generated by a fixed scattering network and \cite{oyallon2019scattering} who show preliminary results using a standard convolutional GAN to generate scattering coefficients. We note that while both approaches are promising, they currently yield suboptimal synthesis results not comparable to modern GANs. Capsule networks \citep{hinton2011transforming,sabour2017dynamic} are also equivariant and emerging work has shown that using a capsule network for the GAN discriminator \citep{capsgan,upadhyay2018generative} improves synthesis on toy datasets. However, capsule GANs and generative scattering approaches require complex training strategies, restrictive architectural choices not compatible with recent insights in GAN training, and have not yet been shown to scale to real-world datasets.

In this work, we improve the generative modeling of images with transformation invariant labels by using an inductive bias of symmetry.
We replace all convolutions with group-convolutions thereby admitting a higher degree of weight sharing which enables increased visual fidelity, especially with limited-sample datasets. 
To our knowledge, we are the first to use group-equivariant layers in the GAN context and to use symmetry-driven considerations in both generator and discriminator architectures.
Our contributions are as follows, 
\begin{enumerate}
    \item We introduce symmetry priors via group-equivariance to generative adversarial networks.
    \item We show that recent insights in improving GAN training are fully compatible with group-equivariance with careful reformulations.
    \item We improve class-conditional image synthesis across a diversity of datasets, architectures, loss functions, and regularizations. These improvements are consistent for both symmetric images and even natural images with preferred orientation.
\end{enumerate}

\section{Methods}

\begin{figure}[!t]
    \centering
    \includegraphics[width=0.85\textwidth]{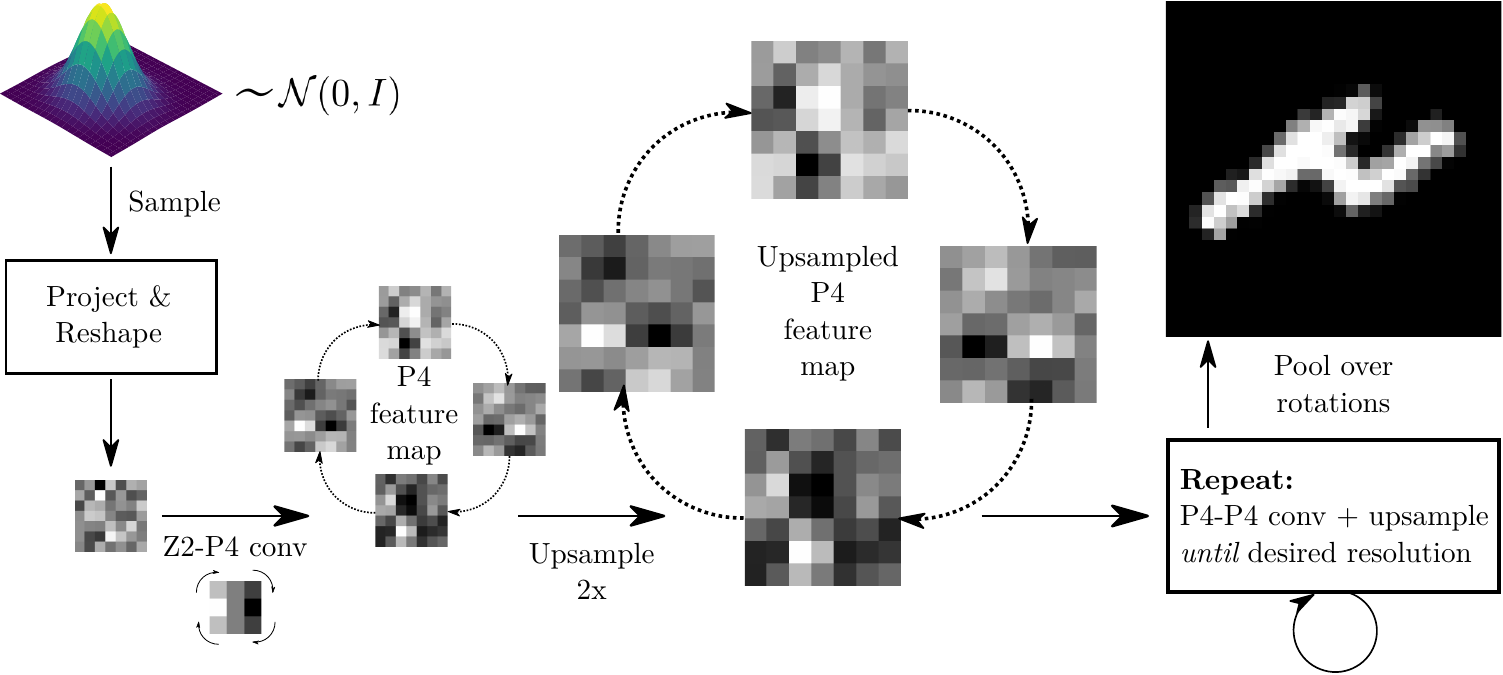}
    \caption{An abbreviated illustration of group-convolutions used in our   generator networks.
    }
    \label{fig:generator}
\end{figure}

\subsection{Preliminaries}

\noindent\textbf{Groups and group-convolutions.} A group is a set with an endowed binary function satisfying the properties of closure, associativity, identity, and invertibility. A two-dimensional symmetry group is the set of all transformations under which a geometric object is invariant with an endowed operation of composition. Given a group $G$ and a map $\Phi: X \rightarrow Y$ between two $G$-sets $X$ and $Y$, $\Phi$ is said to be \textit{equivariant} i.f.f. $\Phi(g \cdot x) = g \cdot \Phi(x), \ \forall x \in X, \ \forall g \in G$. Colloquially, an equivariant map implies that transforming an input and applying the map yields the same result as applying the map and then transforming the output. Analogously, invariance requires that $\Phi(g \cdot x) = \Phi(x), \ \forall x \in X, \ \forall g \in G$. In deep networks, equivariance to a planar symmetry group can be achieved by either transforming filters \citep{cohen2016group} or feature maps \citep{dieleman2016exploiting}.

Our work utilizes the plane symmetry groups $p4$ (all compositions of 90-degree rotations and translations) and $p4m$ (all compositions of 90-degree rotations, reflections, and translations) \citep{schattschneider1978plane}. These groups can be parameterized neatly following \cite{cohen2016group},
\begin{align*}
    g(r, u, v) = 
    \begin{bmatrix}
        \mathrm{cos}(\frac{r\pi}{2}) & -\mathrm{sin}(\frac{r\pi}{2}) & u\\
        \mathrm{sin}(\frac{r\pi}{2}) & \mathrm{cos}(\frac{r\pi}{2}) & v\\
        0 & 0 & 1\\
    \end{bmatrix}; &&
    g'(m, r, u, v) = 
    \begin{bmatrix}
        (-1)^m\mathrm{cos}(\frac{r\pi}{2}) & (-1)^{m+1}\mathrm{sin}(\frac{r\pi}{2}) & u\\
        \mathrm{sin}(\frac{r\pi}{2}) & \mathrm{cos}(\frac{r\pi}{2}) & v\\
        0 & 0 & 1\\
    \end{bmatrix}
\end{align*}
where $g(r, u, v)$ parameterizes $p4$, $g'(m, r, u, v)$ parameterizes $p4m$, $0 \leq r < 4$ (the number of 90-degree rotations), $m \in \{0, 1\}$ (the number of reflections), and $(u, v) \in \mathbb{Z}^2$ (integer translations). The group operation is matrix multiplication for both groups. The matrix $g(r, u, v)$ rotates and translates a point (expressed as homogeneous coordinate vector) in pixel space via left-multiplication. Analogous intuition follows for $g'(m, r, u, v)$.

We now briefly define $G$-equivariant convolutions. We note that formally these are \textit{correlations} and not convolutions and that the literature uses the terms interchangeably. A $G$-convolution between a vector-valued $K$-channel image $f: \mathbb{Z}^2 \rightarrow \mathbb{R}^K$ and filter $\psi: \mathbb{Z}^2 \rightarrow \mathbb{R}^K$ with $f = (f_1, f_2, \dots, f_k)$ and $\psi = (\psi_1, \psi_2, \dots, \psi_k)$ can be expressed as
$
    [f \ast \psi](g) = \sum_{y \in \mathbb{Z}^2} \sum_{k=1}^{K} f_k(y)\psi_{k}(g^{-1}y).
$
For standard reference, if one considers $G$ to be the translation group on $\mathbb{Z}^2$, we have $g^{-1}y = y - g$ and recover the standard convolution. After the first layer of a $G$-CNN, we see that $(f \ast \psi)$ is a function on $G$, necessitating that  filter banks also be functions on $G$. Subsequent $G$-convolutional layers are therefore defined as 
$
    [f \ast \psi](g) = \sum_{h \in G} \sum_{k=1}^{K} f_k(h)\psi_{k}(g^{-1}h).
$
Finally, for tasks where the output is an image, it is necessary to bring the domain of feature maps from $G$ back to $\mathbb{Z}^2$. We can pool the feature map for each filter over the set of transformations, corresponding to average or max pooling over the group of rotations (or roto-reflections as appropriate).

\noindent\textbf{GAN optimization and stability.}
As we focus on the limited data setting where training instability is exacerbated, we briefly describe the two major stabilizing methods used in all experiments here. We regularize the discriminator by using a zero-centered gradient penalty (GP) on the real data as proposed by \citet{mescheder2018training} of the form,
$
    R_1 := \frac{\gamma}{2} \mathbb{E}_{x \sim \mathbb{P}_{real}}[\| \nabla D(x) \|^2_{2}],
$
\noindent where $\gamma$ is the regularization weight, $x$ is sampled from the real distribution $\mathbb{P}_{real}$, and $D$ is the discriminator. This GP has been shown to cause convergence (in toy cases), alleviate catastrophic forgetting \citep{thanh2018catastrophic}, and strongly stabilize GAN training. However, empirical work has found that this GP achieves stability at the cost of worsening GAN evaluation scores \citep{brock2018large}.

A widely used technique for GAN stabilization is spectral normalization \citep{miyato2018spectral}, which constrains the discriminator to be 1-Lipschitz, thereby improving gradient feedback to the generator \citep{pmlr-v97-zhou19c,Chu2020Smoothness}. With spectral normalization, each layer is rescaled as,
$
    W_{SN} = W / \sigma (W),
$
\noindent where $W$ is the weight matrix for a given layer and $\sigma(W)$ is its spectral norm. In practice, $\sigma(W)$ is estimated via a power iteration method as opposed to computing the full singular value decomposition during each training iteration. Finally, applying spectral normalization to both generator and discriminator empirically improves training significantly \citep{zhang2018self}. 

\subsection{Group Equivariant Generative Adversarial Networks} \label{sec:gegans}
Here, we outline how to induce a symmetry prior into the GAN framework. Implementations are available at \url{https://github.com/neel-dey/equivariant-gans}. The literature has developed several techniques for normalization and conditioning of the individual networks, along with unique architectural choices - we extend these developments to the equivariant setting. We start by replacing all convolutional layers with group-convolutional layers where filters and feature maps are functions on a symmetry group $G$. Batch normalization moments \citep{ioffe2015batch} are calculated per group-feature map as opposed to spatial feature maps. Pointwise nonlinearities preserve equivariance for the groups considered here. Pre-activation residual blocks common to modern GANs are used freely as the sum of equivariant feature maps on $G$ is also equivariant.

\noindent\textbf{Generator.} The generator is illustrated at a high-level in Figure \ref{fig:generator}. We use a fully connected layer to linearly project and reshape the concatenated noise vector $z \sim \mathcal{N}(0, I)$ and class embedding $c$ into spatial feature maps on $\mathbb{Z}^2$. We then use spectrally-normalized group-convolutions, interspersed with pointwise-nonlinearities, and nearest-neighbours upsampling to increase spatial extent. We use upsampling followed by group-convolutions instead of transposed group-convolutions to reduce checkerboard artefacts \citep{odena2016deconvolution}. We further use a novel group-equivariant class-conditional batch normalization layer (described below) to normalize and class-condition image generation while also projecting the latent vector $z$ to each level of the group-convolutional hierarchy. We finally max-pool over the set of transformations to obtain the generated image $x$.

\noindent\textbf{Discriminator.} The group-equivariant discriminator receives an input $x$, which it maps to a scalar indicating whether it is real or fake. We do this via spectrally normalized group-convolutions, pointwise-nonlinearities, and spatial-pooling layers to decrease spatial extent. After the final group-convolutional layer, we pool over the group and use global average pooling to obtain an invariant representation at the output. Finally, we condition the discriminator output via the projection method proposed by \citet{miyato2018cgans}. Importantly, the equivariance of group-convolutions depends on the convolutional stride. Strided convolutions were commonly used for downsampling in early GANs \citep{radford2015unsupervised}. However, stride values must be adjusted to the dataset to preserve equivariance, which makes comparisons to equivalent non-equivariant GAN architectures difficult. We therefore use pooling layers over the plane (commonly used in recent GANs) to downsample in all settings to preserve equivariance and enable a fair comparison.

\noindent\textbf{Spectral Normalization.} As the singular values of a matrix are invariant under compositions of 90-degree rotations, transpositions, and reflections - spectral normalization on a group-weight matrix preserves equivariance and we use it freely.

\noindent\textbf{Class-conditional Batch Normalization.} Conditional batch normalization \citep{perez2018film} replaces the scale and shift of features with an affine transformation learned from the class label (and optionally from the latent vector as well \citep{brock2018large}) via linear dense layers, and is widely used in generative networks. We propose a group-equivariance preserving conditional normalization by learning the affine transformation parameters per group-feature map, rather than each spatial feature. As we use fewer group-filters than equivalent non-equivariant GANs, we use fewer dense parameters to learn conditional scales and shifts.

\section{Experiments}

\begin{table}[t]
\begin{center}

\caption{A summary of the datasets considered in this paper. The right-most column indicates whether the dataset has a preferred pose.}
\begin{tabular}{cccccccc}
 \toprule
 \textbf{Dataset} & $\mathbf{Resolution}$ & $\mathbf{n_{classes}}$ & $\mathbf{n_{training}}$ & $\mathbf{n_{validation}}$ & \textbf{Pose Preference} \\
 \midrule
 Rotated MNIST & (28, 28) & 10 & 12,000 & 50,000 & No \\
 ANHIR & (128, 128, 3) & 5 & 28,407 & 9,469 & No \\
 LYSTO & (256, 256, 3) & 3 & 20,000 & - & No \\
 CIFAR-10 & (32, 32, 3) & 10 & 50,000 & 10,000 & Yes \\
 Food-101 & (64, 64, 3) & 101 & 75,747 & 25,250 & Yes \\
 \midrule
\end{tabular}
\label{tab:datasets}

\caption{Min. \& mean Fr\'echet distances (lower is better) of generated RotMNIST samples, evaluated at every 1K generator iterations.  All evaluations are  visualized in Appendix \ref{app:suppl_res} Figure \ref{app:rmnist_res_fig}.
}
\begin{tabular}{cccccc}

 \toprule
 \multicolumn{2}{c}{\textbf{ }} & \multicolumn{4}{c}{Min. \& Mean Fr\'echet Distance}  \\
 \multicolumn{2}{c}{\textbf{ }} & \multicolumn{4}{c}{Available Training Data}  \\
 Loss & Setting & 10\% & 33\% & 66\% & 100\%   \\ 
 \midrule
 - & Real data & 0.6854 & 0.3208 & 0.1324 & 0.1296 \\
 
 \midrule
 
 \multirow{4}{*}{\STAB{\rotatebox[origin=c]{90}{RAGAN}}} & CNN in \verb|G| \& \verb|D|          & (2.04, 11.40) & (1.42, 11.65) & (1.20, 11.10) & (1.36, 11.68) \\ 
                                                         & CNN in \verb|G| \& G-CNN in \verb|D| & (1.84, 4.26)  & (0.88, \textbf{3.26})  & (0.52, 2.85)  & (0.53, 3.12) \\ 
                                                         & G-CNN in \verb|G| \& CNN in \verb|D| & (1.49, 9.75)  & (1.08, 9.29)  & (0.90, 8.70)  & (0.95, 9.62) \\ 
                                                         & G-CNN in \verb|G| \& \verb|D|        & (1.61, \textbf{4.25})  & (0.76, 3.40)  & (0.54, 2.92)  & (0.53, \textbf{2.90}) \\ 
 
 \midrule
 \multirow{4}{*}{\STAB{\rotatebox[origin=c]{90}{NSGAN}}} & CNN in \verb|G| \& \verb|D|          & (\textbf{1.00}, 7.02) & (0.74, 8.25) & (0.84, 8.07) & (0.97, 8.49) \\ 
                                                         & CNN in \verb|G| \& G-CNN in \verb|D| & (2.77, 5.48) & (1.02, 3.51) & (0.55, \textbf{2.85}) & (0.54, 3.08) \\ 
                                                         & G-CNN in \verb|G| \& CNN in \verb|D| & (\textbf{1.00}, 7.00) & (0.96, 7.42) & (0.87, 6.83) & (0.94, 7.52) \\ 
                                                         & G-CNN in \verb|G| \& \verb|D|        & (2.85, 5.67) & (1.04, 4.24) & (0.82, 3.27) & (0.64, 3.32) \\ 
 
 \midrule
 
 \multirow{4}{*}{\STAB{\rotatebox[origin=c]{90}{WGAN}}} & CNN in \verb|G| \& \verb|D|           & (3.42, 16.21) & (3.90, 18.32) & (3.87, 17.81) & (4.88, 19.40) \\ 
                                                         & CNN in \verb|G| \& G-CNN in \verb|D| & (2.87, 5.98)  & (0.76, 4.11)  & (\textbf{0.50}, 3.57)  & (\textbf{0.39}, 3.51) \\ 
                                                         & G-CNN in \verb|G| \& CNN in \verb|D| & (2.67, 16.02) & (3.40, 17.03) & (3.77, 17.76) & (3.74, 17.82) \\ 
                                                         & G-CNN in \verb|G| \& \verb|D|        & (2.51, 5.67)  & (\textbf{0.58}, 3.32)  & (0.56, 3.52)  & (0.54, 3.76) \\ 
 
 \midrule
\end{tabular}
\label{tab:rmnist}

\end{center}
\end{table}

\noindent\textbf{Common setups.} In each subsection, we list specific experimental design choices with full details available in App. \ref{app:exp_det}. For each comparison, the number of group-filters in each layer is divided by the square root of the cardinality of the symmetry set to ensure a similar number of parameters to the standard CNNs to enable fair comparison. We skew towards stabilizing training over absolute performance to compare models under the same settings to obviate extensive checkpointing typically required for BigGAN-like models.
Optimization is performed via Adam \citep{kingma2014adam} with $\beta_1 = 0.0$ and $\beta_2 = 0.9$, as in \citet{zhang2018self,brock2018large}. Unless otherwise noted, all discriminators are updated twice per generator update and employ unequal learning rates for the generator and discriminator following \citet{heusel2017gans}. We use an exponential moving average ($\alpha=0.9999$) of generator weights across iterations when sampling images as in \citet{brock2018large}. All initializations use the same random seed, except for RotMNIST where we average over 3 random seeds. An overview of the small datasets considered here is presented in Table \ref{tab:datasets}.

\textbf{Evaluation methodologies.}  GANs are commonly evaluated by embedding the real and generated images into the feature space of an ImageNet pre-trained network where similarity scores are computed. The Fr\'echet Inception Distance (FID) \citep{heusel2017gans} jointly captures sample fidelity and diversity and is presented for all experiments. To further evaluate both aspects explicitly, we present the improved precision and recall scores  \citep{kynkaanniemi2019improved} for ablations on real-world datasets. As the medical imaging datasets (ANHIR and LYSTO) are not represented in ImageNet, we finetune Inception-v3 \citep{szegedy2016rethinking} prior to feature extraction for FID calculation as in \cite{huang2018multimodal}. For RotMNIST, we use features derived from the final pooling layer of the $p4$-CNN defined in \cite{cohen2016group} to replace Inception-featurization. An analogous approach was taken in \cite{BinkowskiSAG18} in their experiments on the canonical MNIST dataset. Natural image datasets (Food-101 and CIFAR-10) are evaluated with the official \verb|Tensorflow| Inception-v3 weights. Importantly, we perform ablation studies on all datasets to evaluate group-equivariance in either or both networks. 

We note that the FID estimator is strongly biased \citep{BinkowskiSAG18} and work around this limitation by always generating the same number of samples as the validation set as recommended in \cite{BinkowskiSAG18}. An alternative Kernel Inception Distance (KID) with negligible bias has been proposed \citep{BinkowskiSAG18}, yet large-scale evaluation \citep{kurach19gans} finds that KID correlates strongly with FID. We thus focus on FID in our experiments in the main text.

\subsection{Synthetic Experiments: Rotated MNIST} \label{sec:rmnist_results}

Rotated MNIST \citep{larochelle2007empirical} provides random rotations of the MNIST dataset and is a common benchmark for equivariant CNNs which we use to measure sensitivity to dataset size, loss function, and equivariance in either network to motivate choices for real-world experiments.
We experiment with four different proportions of training data: 10\%, 33\%, 66\%, and 100\%. Additionally, the non-saturating loss \citep{goodfellow2014generative} (NSGAN), the Wasserstein loss \citep{arjovsky2017wasserstein} (WGAN), and the relativistic average loss \citep{jolicoeur2018relativistic} (RaGAN) are tested. 
For the equivariant setting, all convolutions are replaced with $p4$-convolutions. $p4m$ is precluded as some digits do not possess mirror symmetry.
All settings were trained for 20,000 generator iterations with a batch size of 64.
Implementation details are available in Appendix \ref{app:rmnist_impl}.

\begin{figure}[t]
    \centering
    \includegraphics[width=0.9\textwidth]{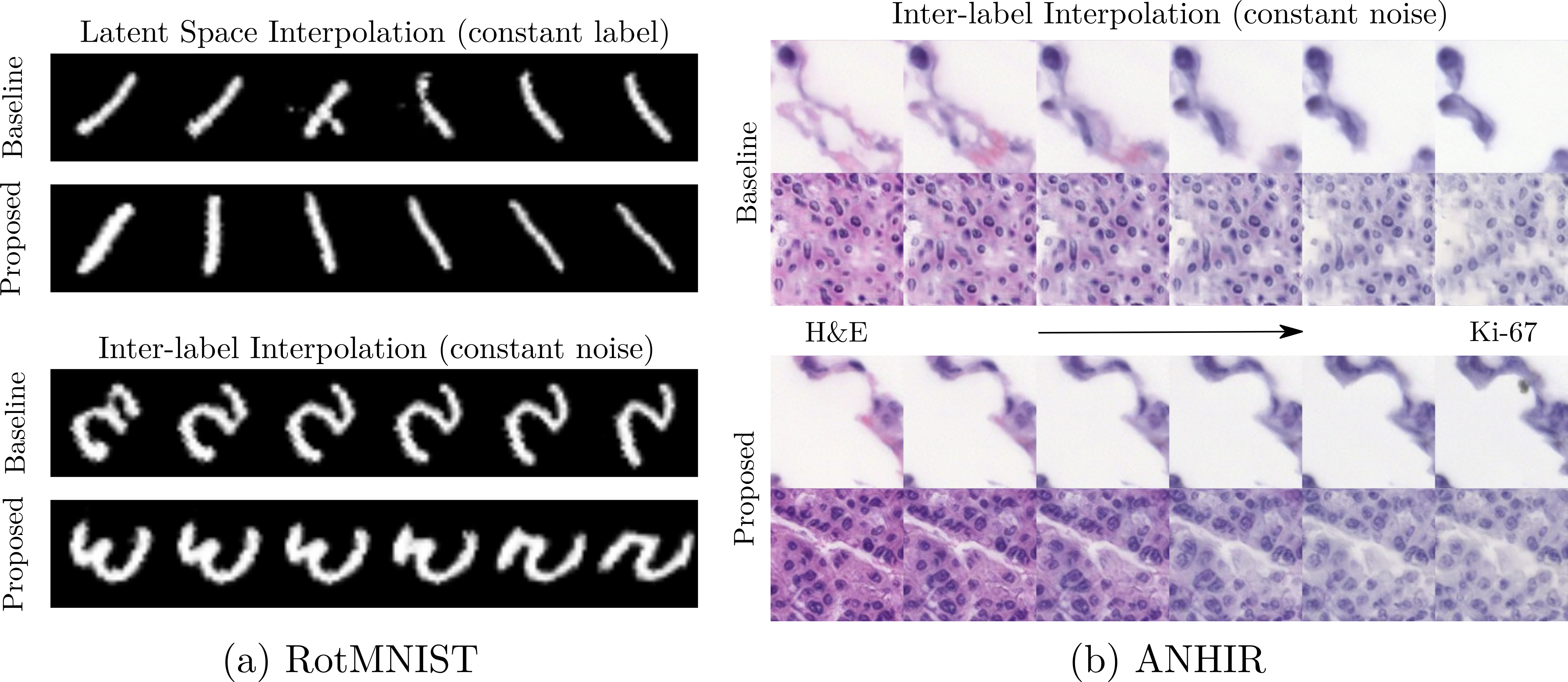}
    \caption{Qualitative GAN interpolation \citep{white2016sampling} results. \textbf{(a)} Selected spherical interpolations between generated RotMNIST samples in either latent space (\textbf{top}) or between labels (\textbf{bottom}). Equivariant GANs interpolate intuitively between samples, whereas standard GANs do not. \textbf{(b)} Selected inter-label linear interpolations between two staining dyes in synthesized ANHIR images. The standard model (\textbf{top}) changes both structure and dye between the generated samples, whereas the equivariant model (\textbf{bottom}) better preserves structure while translating between dyes.
    }
    \label{res:interpolations}
\end{figure}

\noindent\textbf{Results.} Fr\'echet distance of synthesized samples to the validation set is calculated at every thousand generator iterations. As shown in Table \ref{tab:rmnist}, we find that under nearly every configuration of loss and data availability considered, using $p4$-convolutions in either network improves both the mean and minimum Fr\'echet distance. As data availability increases, the best-case minimum and mean FID scores improve.
With $\{33\%, 66\%, 100\%\}$ of the data, most improvements come from using a $p4$-discriminator, with the further usage of a $p4$-generator only helping in a few cases. At $10\%$ data, having an equivariant generator is more impactful than an equivariant discriminator.
These trends are further evident from App. \ref{app:suppl_res} Fig. \ref{app:rmnist_res_fig}, where we see that GANs with $p4$-discriminators converge faster than non-equivariant counterparts.
The NSGAN-GP and RAGAN-GP losses perform similarly, with WGAN-GP underperforming initially and ultimately achieving comparable results.
Qualitatively, the equivariant model learns better representations as shown in Figure \ref{res:interpolations}(a). Holding the class-label constant and interpolating between samples, we find that the standard GAN changes the shape of the digit in order to rotate it, whereas the equivariant model learns rotation in the latent space. Holding the latent constant and interpolating between classes shows that our model learns an intuitive interpolation between digits, whereas the standard GAN transforms the image immediately.

\begin{table}[t]
  \centering
  \caption{FID evaluation (lower is better) of all real-world datasets across ablations and augmentation-based baseline comparisons. - indicates an inapplicable setting for the method.}
  \begin{tabular}{clcccc}
    \hline
     & \multicolumn{1}{c}{Setting} & \multicolumn{1}{c}{ANHIR} & \multicolumn{1}{c}{LYSTO} & \multicolumn{1}{c}{CIFAR-10} & \multicolumn{1}{c}{Food-101} \\ \hline
     \multirow{4}{*}{\STAB{\rotatebox[origin=c]{90}{Ablation}}}
     & CNN in \verb|G| \& \verb|D|  & 7.32 & 7.27 & 20.89 & 27.34\\
     & G-CNN in \verb|G|; CNN in \verb|D| & 6.93 & 6.68 & 21.20 & 24.16 \\
     & CNN in \verb|G|; G-CNN in \verb|D| & 5.56 & 5.02 & \textbf{17.09} & \textbf{16.91} \\
     & G-CNN in \verb|G| \& \verb|D| & \textbf{5.54} & \textbf{3.90} & 17.49 & 17.73 \\ \hline
     \multirow{4}{*}{\STAB{\rotatebox[origin=c]{90}{+ Aug.}}}
     & CNN in \verb|G| \& \verb|D| + Standard Aug. & 7.57 & 6.59 & 37.41 & 35.18 \\
     & CNN in \verb|G| \& \verb|D| + bCR \citep{zhao2020improved} & 5.86 & 4.78 & 19.64  & 21.18 \\
     & CNN in \verb|G| \& \verb|D| + AR \citep{chen2019self} & - & - & 19.59 & 20.39 \\
     & G-CNN in \verb|G| \& \verb|D| + bCR \citep{zhao2020improved} & \textbf{5.19} & \textbf{4.53} & \textbf{17.94} & \textbf{15.55} \\ \hline
  \end{tabular}
  \label{tab:real_quant_results}
\end{table}

\subsection{Real-world Experiments} \label{sec:realexp}

\textbf{Datasets.} $p4$ and $p4m$-equivariant networks are most useful when datasets possess global roto(-reflective) symmetry, yet have also been shown to benefit generic image representation due to local symmetries \citep{cohen2016group,romero2020att}. To this end, we experiment with two types of real-world  datasets as detailed in Table \ref{tab:datasets}: (1) sets with roto(-reflective) symmetry, such that the image label is invariant under transformation; (2) natural images with preferred orientation (e.g., the \verb|boat| class of images in CIFAR-10 cannot be upside-down). Briefly, they are:

\textit{ANHIR} provides high-resolution pathology slides stained with 5 different dyes to highlight different cellular properties \citep{ANHIR,borovec2018benchmarking}. We extract $128 \times 128$ foreground patches from images of different scales, as described in App. \ref{app:anhir_patch}. We use the staining dye as conditioning.

\textit{LYSTO} is a multi-organ pathology benchmark for the counting of immunohistochemistry stained lymphocytes \citep{francesco_ciompi_2019_3513571}. We re-purpose it here for conditional synthesis at a higher resolution of $256 \times 256$. As classification labels are not provided, we use the organ source as class labels. The use of organ sources as classes is validated in App. \ref{app:lysto_organ}. The high image resolution in addition to the limited sample size of 20,000 make LYSTO a challenging dataset for GANs.

\textit{CIFAR-10} is a natural image vision benchmark of both small resolution and sample size \citep{krizhevsky2009learning}. Previous work \citep{weiler2019general,romero2020att} finds that equivariant-networks improve classification accuracy on CIFAR-10 and we include here it as a GAN benchmark.

\textit{Food-101} is a small natural image dataset of a 101 categories of food taken in various challenging settings of over/under exposure, label noise, etc. \citep{bossard14}. Further, datasets with a high number of classes are known to be challenging for GANs \citep{odena2019open}. Importantly, even though the objects in this dataset have a preferred pose due to common camera orientations, we speculate that roto-equivariance may be beneficial here as food photography commonly takes an \textit{en face} or mildly oblique view. We resize the training set to $64 \times 64$ resolution for our experiments.

\begin{figure}[t]
    \centering
    \includegraphics[width=1\textwidth]{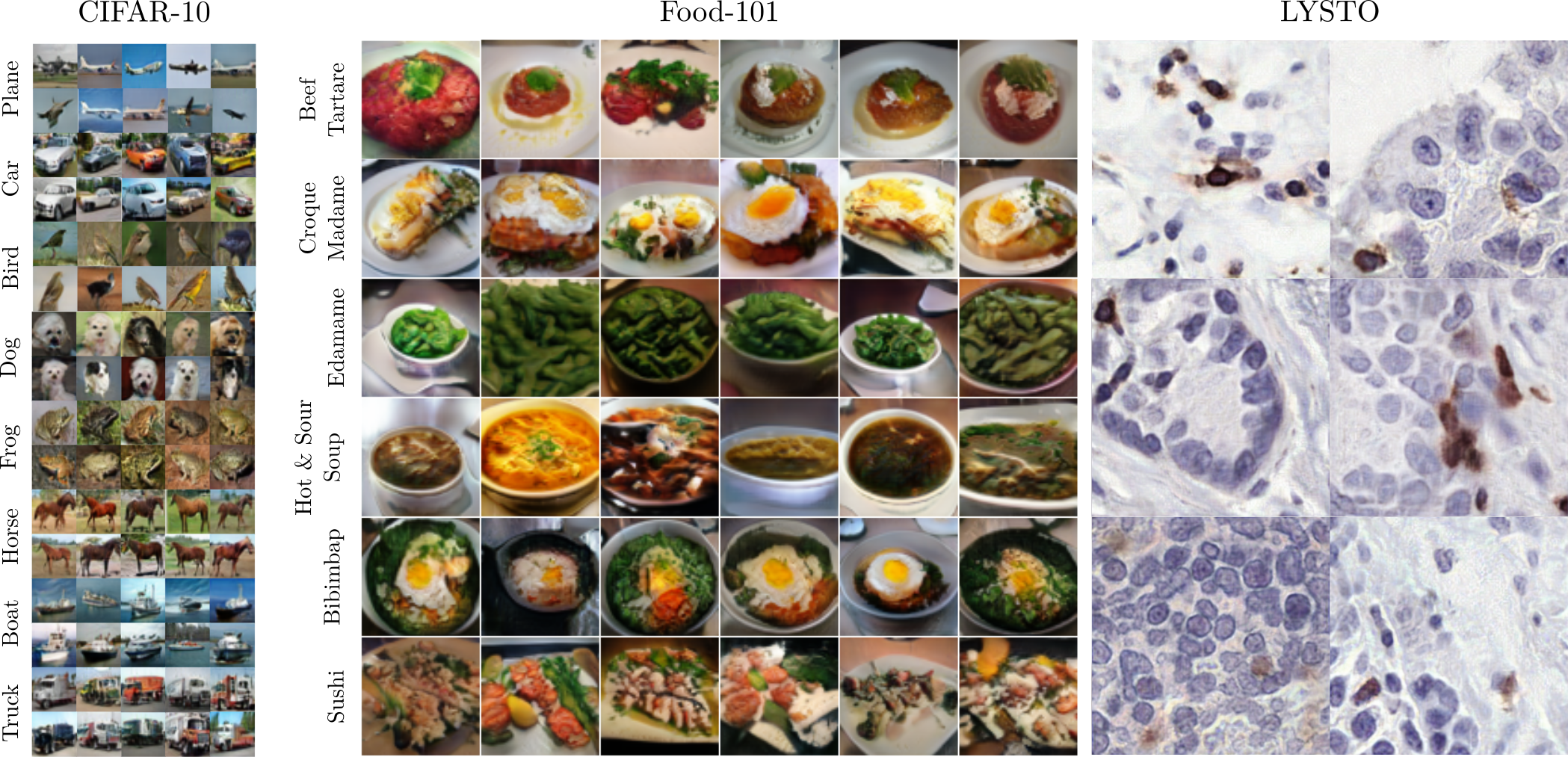}
    \caption{Selected generated samples using the best performing equivariant models with no augmentation. Random samples are available in App. \ref{app:suppl_res}. Layout inspired by \citet{karras2020training}.}
    \label{res:gen_collage}
\end{figure}

\textbf{Baseline architecture.} To produce a strong non-equivariant baseline, we face several design choices. State-of-the-art GANs follow either BigGAN \citep{brock2018large} or StyleGAN2 \citep{Karras_2020_CVPR} in design. As StyleGAN2 has not yet been demonstrated to scale to \textit{conditional} generation with a large number of classes (to our knowledge), we follow a BigGAN-like construction despite the stability of StyleGAN2. For our small datasets, we make the following modifications: (1) we use fewer channels; (2) we do not use orthogonal regularization; (3) we do not use hierarchical latent projection as we find in early testing that projecting the entire latent to each normalization layer achieves similar results; (4) we do not use attention as equivariant attention is an area of active research \citep{romero2019co,romero2020att} but currently has prohibitively high memory requirements and may not yet scale to GANs. Further details are available in App. \ref{app:addn_impl_details}.

We then modify either generator (\verb|G|) and/or discriminator (\verb|D|) as in Section \ref{sec:gegans} to obtain the corresponding equivariant settings.
We note that a discriminator invariant to roto-reflections would assign the same amount of realism to an upright natural image versus a rotated/reflected copy of the same image, allowing the generator to synthesize images at arbitrary orientations. 
Therefore, for CIFAR-10 and Food-101 we pool over rotations \textit{before} the last residual block  to enable the discriminator to detect when generated images are not in their canonical pose while maintaining most of the benefits of equivariance as studied in \citet{weiler2019general}. We use $p4m$-equivariance for ANHIR and LYSTO and $p4$-equivariance for CIFAR-10 and Food-101 to reduce training time.

\begin{figure}[t]
    \centering
    \includegraphics[width=1\textwidth]{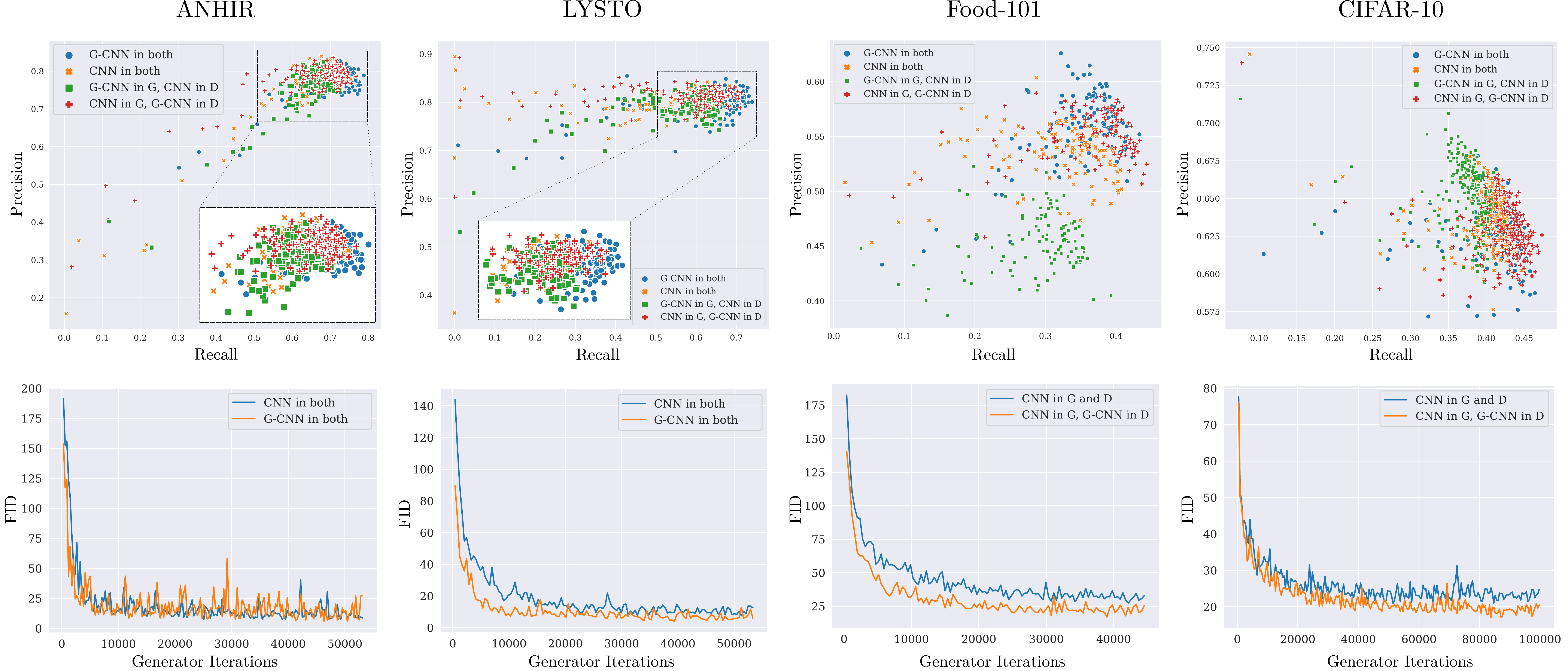}
    \caption{\textbf{Top:} Improved Precision and Recall \citep{kynkaanniemi2019improved} analysis of ablations for all snapshots of trained models in each setting (closer to top-right is better). \textbf{Bottom:} GAN convergence (FID vs. generator updates) of standard GANs vs. our proposed models for all datasets. For visual clarity, we show only a subset of comparisons with convergence plots for all methods provided in App. \ref{app:suppl_res} Fig. \ref{app:all_convergence_plots}. Readers are encouraged to zoom-in for better inspection.}
    \label{res:ablation_pr_collage}
\end{figure}

\textbf{Comparisons.} A natural comparison would be against standard GANs using augmentations drawn from the same group our model is equivariant to. However, augmentation on the real images alone would lead to the augmentations ``leaking" into the generated images, e.g., vertical flip augmentation may lead to generated images being upside-down. \citet{zhao2020improved} propose balanced consistency regularization (bCR) for augmentations of both real and generated samples to alleviate this issue, and we thus use it as a comparison. We restrict the augmentations used in bCR to
90-degree rotations or 90-degree rotations and reflections
as appropriate to enable a fair comparison against equivariant GANs. Using additional augmentations would help all methods across the board. We further compare against auxiliary rotations (AR) GAN \citep{chen2019self} where real and fake images are augmented with 90-degree rotations and the discriminator is tasked with predicting their orientation. We do not use AR for ANHIR and LYSTO as they have no canonical orientation. For completeness, we also evaluate standard augmentation (reals only) for all datasets.

\textbf{Results.} Quantitative FID results of ablations and comparisons against baselines are presented in Table \ref{tab:real_quant_results}. Equivariant networks (G-CNNs) outperform methods which use standard CNNs with or without augmentation across all datasets. For ANHIR and LYSTO, we find that $p4m$-equivariance in either network improves FID evaluation, with the best results coming from modifying both networks. However, for the upright datasets CIFAR-10 and Food-101, we find that having a $p4$-equivariant discriminator alone helps more than having both networks be $p4$-equivariant. We speculate that this effect is in part attributable to their orientation bias. With bCR and AR GANs, we find that standard CNNs improve significantly, yet are still outperformed by equivariant nets using no augmentation. We include a mixture of equivariant GANs and bCR for completeness and find that for ANHIR and Food-101, they have an additive effect, whereas they do not for LYSTO and CIFAR-10, indicating a dataset-sensitivity. Of note, we found that bCR with its suggested hyperparameters lead to immediate training collapse on ANHIR, LYSTO, and CIFAR-10, which was fixed by decreasing the strength of the regularization substantially. This may be due to the original work using several different types of augmentation and not just roto-reflections. Standard augmentation (i.e., augmenting training images alone) lead to augmentation leakage for CIFAR-10 and Food-101.

Qualitatively, as class differences in ANHIR should be stain (color) based, we visualize inter-class interpolations between synthesized samples in Figure \ref{res:interpolations}(b). We find that our model better preserves structure while translating between stains, whereas the non-equivariant GAN struggles to do so. In our ablation study in terms of precision and recall in Figure \ref{res:ablation_pr_collage}, using $p4m$-equivariance in \verb|G| and \verb|D| achieves consistently higher recall for ANHIR and LYSTO. For Food-101, we find that \verb|G-CNN in G and D| achieves higher precision, whereas \verb|CNN in G and G-CNN in D| achieves higher recall.
For CIFAR-10 precision and recall, we find no discernable differences between the two settings with lowest FID. 
Interestingly, for CIFAR-10 adding $p4$-equivariance to \verb|G| but not \verb|D| worsens FID but noticeably improves precision. These observations are consistent with our FID findings as FID tends to correlate better with recall \citep{karras2020training}. Finally, we plot FID vs. generator updates in Figure \ref{res:ablation_pr_collage}, finding that the proposed framework converges faster than the baseline as a function of training iterations (for all datasets except ANHIR). Convergence plots for all datasets and all methods compared can be found in App. \ref{app:suppl_res} Figure \ref{app:all_convergence_plots}, showing similar trends.

\section{Discussion}
\textbf{Future work.} 
We present improved conditional image synthesis using equivariant networks, opening several potential future research directions:
(1) As efficient implementations of equivariant attention develop, we will incorporate them to model long-range dependency;
(2) Equivariance to continuous groups may yield further increased data efficiency and more powerful representations. However, doing so may require non-trivial modifications to current GAN architectures as memory limitations could bottleneck continuous group-equivariant GANs at relevant image sizes. Further, adding more discretizations beyond 4 rotations on a continuous group such has $SE(2)$ may show diminishing returns \citep[~Fig.7]{lafarge2020rototranslation}; 
(3) In parallel to our work, \citet{karras2020training} propose a differentiable augmentation scheme for limited data GANs pertaining to \textit{which} transformations to apply and learning the frequency of augmentation for generic images, with similar work presented in \citet{zhao2020differentiable}. Our approach is fully complementary to these methods when employing transformations outside the considered group and will be integrated into future work;
(4) Contemporaneously, \citet{lafarge2020orientationdisentangled} propose equivariant variational autoencoders allowing for control over generated orientations via structured latent spaces which may be used for equivariant GANs as well; (5) The groups considered here do not capture all variabilities present in natural images such as small diffeomorphic warps. Scattering networks may provide an elegant framework to construct GANs equivariant to a wider range of symmetries and enable higher data-efficiency.

\textbf{Conclusion.} We present a flexible framework for incorporating symmetry priors within GANs. In doing so, we improve the visual fidelity of GANs in the limited-data regime when trained on symmetric images and even extending to natural images. Our experiments confirm this by improving on conventional GANs across a variety of datasets, ranging from medical imaging modalities to real-world images of food. Modifying either generator or discriminator generally leads to improvements in synthesis, with the latter typically having more impact. To our knowledge, our work is the first to show clear benefits of equivariant learning over standard GAN training on high-resolution conditional image generation beyond toy datasets. While this work is empirical, we believe that it strongly motivates future theoretical analysis of the interplay between GANs and equivariance. Finally, improved results over augmentation-based strategies are presented, demonstrating the benefits of explicit transformation equivariance over equivariance-approximating regularizations.

\textbf{Acknowledgements.} Neel Dey thanks Mengwei Ren, Axel Elaldi, Jorge Ono, and Guido Gerig.

\bibliography{references}

\begin{thebibliography}{74}
\providecommand{\natexlab}[1]{#1}
\providecommand{\url}[1]{\texttt{#1}}
\expandafter\ifx\csname urlstyle\endcsname\relax
  \providecommand{\doi}[1]{doi: #1}\else
  \providecommand{\doi}{doi: \begingroup \urlstyle{rm}\Url}\fi

\bibitem[Angles \& Mallat(2018)Angles and Mallat]{angles2018generative}
Tomás Angles and Stéphane Mallat.
\newblock Generative networks as inverse problems with scattering transforms.
\newblock In \emph{International Conference on Learning Representations}, 2018.
\newblock URL \url{https://openreview.net/forum?id=r1NYjfbR-}.

\bibitem[Arjovsky et~al.(2017)Arjovsky, Chintala, and
  Bottou]{arjovsky2017wasserstein}
Martin Arjovsky, Soumith Chintala, and L{\'e}on Bottou.
\newblock Wasserstein gan.
\newblock \emph{arXiv preprint arXiv:1701.07875}, 2017.

\bibitem[Bietti \& Mairal(2019)Bietti and Mairal]{bietti2019group}
Alberto Bietti and Julien Mairal.
\newblock Group invariance, stability to deformations, and complexity of deep
  convolutional representations.
\newblock \emph{The Journal of Machine Learning Research}, 20\penalty0
  (1):\penalty0 876--924, 2019.

\bibitem[Binkowski et~al.(2018)Binkowski, Sutherland, Arbel, and
  Gretton]{BinkowskiSAG18}
Mikolaj Binkowski, Dougal~J. Sutherland, Michael Arbel, and Arthur Gretton.
\newblock Demystifying {MMD} {GAN}s.
\newblock In \emph{6th International Conference on Learning Representations,
  {ICLR} 2018, Vancouver, BC, Canada, April 30 - May 3, 2018, Conference Track
  Proceedings}, 2018.

\bibitem[Borovec et~al.(2018)Borovec, Munoz-Barrutia, and
  Kybic]{borovec2018benchmarking}
Ji{\v{r}}{\'\i} Borovec, Arrate Munoz-Barrutia, and Jan Kybic.
\newblock Benchmarking of image registration methods for differently stained
  histological slides.
\newblock In \emph{2018 25th IEEE International Conference on Image Processing
  (ICIP)}, pp.\  3368--3372. IEEE, 2018.

\bibitem[Borovec et~al.(2020)Borovec, Kybic, Arganda-Carreras, Sorokin, Bueno,
  Khvostikov, Bakas, Eric, Chang, Heldmann, et~al.]{ANHIR}
Ji{\v{r}}{\'\i} Borovec, Jan Kybic, Ignacio Arganda-Carreras, Dmitry~V Sorokin,
  Gloria Bueno, Alexander~V Khvostikov, Spyridon Bakas, I~Eric, Chao Chang,
  Stefan Heldmann, et~al.
\newblock Anhir: automatic non-rigid histological image registration challenge.
\newblock \emph{IEEE Transactions on Medical Imaging}, 2020.

\bibitem[Bossard et~al.(2014)Bossard, Guillaumin, and Van~Gool]{bossard14}
Lukas Bossard, Matthieu Guillaumin, and Luc Van~Gool.
\newblock Food-101 -- mining discriminative components with random forests.
\newblock In \emph{European Conference on Computer Vision}, 2014.

\bibitem[Brock et~al.(2018)Brock, Donahue, and Simonyan]{brock2018large}
Andrew Brock, Jeff Donahue, and Karen Simonyan.
\newblock Large scale gan training for high fidelity natural image synthesis.
\newblock \emph{arXiv preprint arXiv:1809.11096}, 2018.

\bibitem[{Bruna} \& {Mallat}(2013){Bruna} and {Mallat}]{bruna2013scattering}
J.~{Bruna} and S.~{Mallat}.
\newblock Invariant scattering convolution networks.
\newblock \emph{IEEE Transactions on Pattern Analysis and Machine
  Intelligence}, 35\penalty0 (8):\penalty0 1872--1886, 2013.
\newblock \doi{10.1109/TPAMI.2012.230}.

\bibitem[Chen et~al.(2019)Chen, Zhai, Ritter, Lucic, and Houlsby]{chen2019self}
Ting Chen, Xiaohua Zhai, Marvin Ritter, Mario Lucic, and Neil Houlsby.
\newblock Self-supervised gans via auxiliary rotation loss.
\newblock In \emph{Proceedings of the IEEE Conference on Computer Vision and
  Pattern Recognition}, pp.\  12154--12163, 2019.

\bibitem[Chidester et~al.(2019)Chidester, Ton, Tran, Ma, and
  Do]{chidester2019enhanced}
Benjamin Chidester, That-Vinh Ton, Minh-Triet Tran, Jian Ma, and Minh~N Do.
\newblock Enhanced rotation-equivariant u-net for nuclear segmentation.
\newblock In \emph{Proceedings of the IEEE Conference on Computer Vision and
  Pattern Recognition Workshops}, pp.\  0--0, 2019.

\bibitem[Chu et~al.(2020)Chu, Minami, and Fukumizu]{Chu2020Smoothness}
Casey Chu, Kentaro Minami, and Kenji Fukumizu.
\newblock Smoothness and stability in gans.
\newblock In \emph{International Conference on Learning Representations}, 2020.
\newblock URL \url{https://openreview.net/forum?id=HJeOekHKwr}.

\bibitem[Ciompi et~al.(2019)Ciompi, Jiao, and van~der
  Laak]{francesco_ciompi_2019_3513571}
Francesco Ciompi, Yiping Jiao, and Jeroen van~der Laak.
\newblock Lymphocyte assessment hackathon (lysto), October 2019.
\newblock URL \url{https://doi.org/10.5281/zenodo.3513571}.

\bibitem[Cohen \& Welling(2016)Cohen and Welling]{cohen2016group}
Taco Cohen and Max Welling.
\newblock Group equivariant convolutional networks.
\newblock In \emph{International conference on machine learning}, pp.\
  2990--2999, 2016.

\bibitem[Cohen et~al.(2019)Cohen, Geiger, and Weiler]{cohen2019general}
Taco~S Cohen, Mario Geiger, and Maurice Weiler.
\newblock A general theory of equivariant cnns on homogeneous spaces.
\newblock In \emph{Advances in Neural Information Processing Systems}, pp.\
  9142--9153, 2019.

\bibitem[Dieleman et~al.(2016)Dieleman, De~Fauw, and
  Kavukcuoglu]{dieleman2016exploiting}
Sander Dieleman, Jeffrey De~Fauw, and Koray Kavukcuoglu.
\newblock Exploiting cyclic symmetry in convolutional neural networks.
\newblock \emph{arXiv preprint arXiv:1602.02660}, 2016.

\bibitem[Dieng et~al.(2019)Dieng, Ruiz, Blei, and Titsias]{dieng2019prescribed}
Adji~B Dieng, Francisco~JR Ruiz, David~M Blei, and Michalis~K Titsias.
\newblock Prescribed generative adversarial networks.
\newblock \emph{arXiv preprint arXiv:1910.04302}, 2019.

\bibitem[Esteves(2020)]{esteves2020theoretical}
Carlos Esteves.
\newblock Theoretical aspects of group equivariant neural networks.
\newblock \emph{arXiv preprint arXiv:2004.05154}, 2020.

\bibitem[Goodfellow et~al.(2014)Goodfellow, Pouget-Abadie, Mirza, Xu,
  Warde-Farley, Ozair, Courville, and Bengio]{goodfellow2014generative}
Ian Goodfellow, Jean Pouget-Abadie, Mehdi Mirza, Bing Xu, David Warde-Farley,
  Sherjil Ozair, Aaron Courville, and Yoshua Bengio.
\newblock Generative adversarial nets.
\newblock In \emph{Advances in neural information processing systems}, pp.\
  2672--2680, 2014.

\bibitem[Gulrajani et~al.(2017)Gulrajani, Ahmed, Arjovsky, Dumoulin, and
  Courville]{gulrajani2017improved}
Ishaan Gulrajani, Faruk Ahmed, Martin Arjovsky, Vincent Dumoulin, and Aaron~C
  Courville.
\newblock Improved training of wasserstein gans.
\newblock In \emph{Advances in neural information processing systems}, pp.\
  5767--5777, 2017.

\bibitem[Heusel et~al.(2017)Heusel, Ramsauer, Unterthiner, Nessler, and
  Hochreiter]{heusel2017gans}
Martin Heusel, Hubert Ramsauer, Thomas Unterthiner, Bernhard Nessler, and Sepp
  Hochreiter.
\newblock Gans trained by a two time-scale update rule converge to a local nash
  equilibrium.
\newblock In \emph{Advances in Neural Information Processing Systems}, pp.\
  6626--6637, 2017.

\bibitem[Hinton et~al.(2011)Hinton, Krizhevsky, and
  Wang]{hinton2011transforming}
Geoffrey~E Hinton, Alex Krizhevsky, and Sida~D Wang.
\newblock Transforming auto-encoders.
\newblock In \emph{International Conference on Artificial Neural Networks},
  pp.\  44--51. Springer, 2011.

\bibitem[Huang et~al.(2018)Huang, Liu, Belongie, and
  Kautz]{huang2018multimodal}
Xun Huang, Ming-Yu Liu, Serge Belongie, and Jan Kautz.
\newblock Multimodal unsupervised image-to-image translation.
\newblock In \emph{European Conference on Computer Vision}, pp.\  179--196.
  Springer, 2018.

\bibitem[Ioffe \& Szegedy(2015)Ioffe and Szegedy]{ioffe2015batch}
Sergey Ioffe and Christian Szegedy.
\newblock Batch normalization: Accelerating deep network training by reducing
  internal covariate shift.
\newblock \emph{arXiv preprint arXiv:1502.03167}, 2015.

\bibitem[Isola et~al.(2017)Isola, Zhu, Zhou, and Efros]{isola2017image}
Phillip Isola, Jun-Yan Zhu, Tinghui Zhou, and Alexei~A Efros.
\newblock Image-to-image translation with conditional adversarial networks.
\newblock In \emph{Proceedings of the IEEE conference on computer vision and
  pattern recognition}, pp.\  1125--1134, 2017.

\bibitem[Jaiswal et~al.(2019)Jaiswal, AbdAlmageed, Wu, and Natarajan]{capsgan}
Ayush Jaiswal, Wael AbdAlmageed, Yue Wu, and Premkumar Natarajan.
\newblock Capsulegan: Generative adversarial capsule network.
\newblock In Laura Leal-Taix{\'e} and Stefan Roth (eds.), \emph{Computer Vision
  -- ECCV 2018 Workshops}, Cham, 2019. Springer International Publishing.

\bibitem[Jolicoeur-Martineau(2018)]{jolicoeur2018relativistic}
Alexia Jolicoeur-Martineau.
\newblock The relativistic discriminator: a key element missing from standard
  gan.
\newblock \emph{arXiv preprint arXiv:1807.00734}, 2018.

\bibitem[Karras et~al.(2017)Karras, Aila, Laine, and
  Lehtinen]{karras2017progressive}
Tero Karras, Timo Aila, Samuli Laine, and Jaakko Lehtinen.
\newblock Progressive growing of gans for improved quality, stability, and
  variation.
\newblock \emph{arXiv preprint arXiv:1710.10196}, 2017.

\bibitem[Karras et~al.(2020{\natexlab{a}})Karras, Aittala, Hellsten, Laine,
  Lehtinen, and Aila]{karras2020training}
Tero Karras, Miika Aittala, Janne Hellsten, Samuli Laine, Jaakko Lehtinen, and
  Timo Aila.
\newblock Training generative adversarial networks with limited data,
  2020{\natexlab{a}}.

\bibitem[Karras et~al.(2020{\natexlab{b}})Karras, Laine, Aittala, Hellsten,
  Lehtinen, and Aila]{Karras_2020_CVPR}
Tero Karras, Samuli Laine, Miika Aittala, Janne Hellsten, Jaakko Lehtinen, and
  Timo Aila.
\newblock Analyzing and improving the image quality of stylegan.
\newblock In \emph{Proceedings of the IEEE/CVF Conference on Computer Vision
  and Pattern Recognition (CVPR)}, June 2020{\natexlab{b}}.

\bibitem[Kim et~al.(2020)Kim, Kim, Kang, and Lee]{kim2019u}
Junho Kim, Minjae Kim, Hyeonwoo Kang, and Kwang~Hee Lee.
\newblock U-gat-it: Unsupervised generative attentional networks with adaptive
  layer-instance normalization for image-to-image translation.
\newblock In \emph{International Conference on Learning Representations}, 2020.

\bibitem[Kingma \& Ba(2014)Kingma and Ba]{kingma2014adam}
Diederik~P Kingma and Jimmy Ba.
\newblock Adam: A method for stochastic optimization.
\newblock \emph{arXiv preprint arXiv:1412.6980}, 2014.

\bibitem[Krizhevsky et~al.(2009)]{krizhevsky2009learning}
Alex Krizhevsky et~al.
\newblock Learning multiple layers of features from tiny images.
\newblock 2009.

\bibitem[Kurach et~al.(2019)Kurach, Lu{\v{c}}i{\'c}, Zhai, Michalski, and
  Gelly]{kurach19gans}
Karol Kurach, Mario Lu{\v{c}}i{\'c}, Xiaohua Zhai, Marcin Michalski, and
  Sylvain Gelly.
\newblock A large-scale study on regularization and normalization in {GAN}s.
\newblock In Kamalika Chaudhuri and Ruslan Salakhutdinov (eds.),
  \emph{Proceedings of the 36th International Conference on Machine Learning},
  volume~97 of \emph{Proceedings of Machine Learning Research}, pp.\
  3581--3590, Long Beach, California, USA, 09--15 Jun 2019. PMLR.

\bibitem[Kynk{\"a}{\"a}nniemi et~al.(2019)Kynk{\"a}{\"a}nniemi, Karras, Laine,
  Lehtinen, and Aila]{kynkaanniemi2019improved}
Tuomas Kynk{\"a}{\"a}nniemi, Tero Karras, Samuli Laine, Jaakko Lehtinen, and
  Timo Aila.
\newblock Improved precision and recall metric for assessing generative models.
\newblock In \emph{Advances in Neural Information Processing Systems}, pp.\
  3929--3938, 2019.

\bibitem[Lafarge et~al.(2020{\natexlab{a}})Lafarge, Bekkers, Pluim, Duits, and
  Veta]{lafarge2020rototranslation}
Maxime~W. Lafarge, Erik~J. Bekkers, Josien P.~W. Pluim, Remco Duits, and Mitko
  Veta.
\newblock Roto-translation equivariant convolutional networks: Application to
  histopathology image analysis, 2020{\natexlab{a}}.

\bibitem[Lafarge et~al.(2020{\natexlab{b}})Lafarge, Pluim, and
  Veta]{lafarge2020orientationdisentangled}
Maxime~W. Lafarge, Josien P.~W. Pluim, and Mitko Veta.
\newblock Orientation-disentangled unsupervised representation learning for
  computational pathology, 2020{\natexlab{b}}.

\bibitem[Larochelle et~al.(2007)Larochelle, Erhan, Courville, Bergstra, and
  Bengio]{larochelle2007empirical}
Hugo Larochelle, Dumitru Erhan, Aaron Courville, James Bergstra, and Yoshua
  Bengio.
\newblock An empirical evaluation of deep architectures on problems with many
  factors of variation.
\newblock In \emph{Proceedings of the 24th international conference on Machine
  learning}, pp.\  473--480. ACM, 2007.

\bibitem[Lim \& Ye(2017)Lim and Ye]{lim2017geometric}
Jae~Hyun Lim and Jong~Chul Ye.
\newblock Geometric gan.
\newblock \emph{arXiv preprint arXiv:1705.02894}, 2017.

\bibitem[Mallat(2012)]{mallat2012scattering}
Stéphane Mallat.
\newblock Group invariant scattering.
\newblock \emph{Communications on Pure and Applied Mathematics}, 65\penalty0
  (10):\penalty0 1331--1398, 2012.
\newblock \doi{https://doi.org/10.1002/cpa.21413}.
\newblock URL \url{https://onlinelibrary.wiley.com/doi/abs/10.1002/cpa.21413}.

\bibitem[Mescheder et~al.(2018)Mescheder, Geiger, and
  Nowozin]{mescheder2018training}
Lars Mescheder, Andreas Geiger, and Sebastian Nowozin.
\newblock Which training methods for gans do actually converge?
\newblock \emph{arXiv preprint arXiv:1801.04406}, 2018.

\bibitem[Miyato \& Koyama(2018)Miyato and Koyama]{miyato2018cgans}
Takeru Miyato and Masanori Koyama.
\newblock cgans with projection discriminator.
\newblock \emph{arXiv preprint arXiv:1802.05637}, 2018.

\bibitem[Miyato et~al.(2018)Miyato, Kataoka, Koyama, and
  Yoshida]{miyato2018spectral}
Takeru Miyato, Toshiki Kataoka, Masanori Koyama, and Yuichi Yoshida.
\newblock Spectral normalization for generative adversarial networks.
\newblock \emph{arXiv preprint arXiv:1802.05957}, 2018.

\bibitem[Mo et~al.(2020)Mo, Cho, and Shin]{mo2020freeze}
Sangwoo Mo, Minsu Cho, and Jinwoo Shin.
\newblock Freeze the discriminator: a simple baseline for fine-tuning gans,
  2020.

\bibitem[Noguchi \& Harada(2019)Noguchi and Harada]{noguchi2019image}
Atsuhiro Noguchi and Tatsuya Harada.
\newblock Image generation from small datasets via batch statistics adaptation.
\newblock \emph{arXiv preprint arXiv:1904.01774}, 2019.

\bibitem[Odena(2019)]{odena2019open}
Augustus Odena.
\newblock Open questions about generative adversarial networks.
\newblock \emph{Distill}, 2019.
\newblock \doi{10.23915/distill.00018}.
\newblock https://distill.pub/2019/gan-open-problems.

\bibitem[Odena et~al.(2016)Odena, Dumoulin, and Olah]{odena2016deconvolution}
Augustus Odena, Vincent Dumoulin, and Chris Olah.
\newblock Deconvolution and checkerboard artifacts.
\newblock \emph{Distill}, 1\penalty0 (10):\penalty0 e3, 2016.

\bibitem[{Oyallon} et~al.(2019){Oyallon}, {Zagoruyko}, {Huang}, {Komodakis},
  {Lacoste-Julien}, {Blaschko}, and {Belilovsky}]{oyallon2019scattering}
E.~{Oyallon}, S.~{Zagoruyko}, G.~{Huang}, N.~{Komodakis}, S.~{Lacoste-Julien},
  M.~{Blaschko}, and E.~{Belilovsky}.
\newblock Scattering networks for hybrid representation learning.
\newblock \emph{IEEE Transactions on Pattern Analysis and Machine
  Intelligence}, 41\penalty0 (9):\penalty0 2208--2221, 2019.
\newblock \doi{10.1109/TPAMI.2018.2855738}.

\bibitem[Perez et~al.(2018)Perez, Strub, De~Vries, Dumoulin, and
  Courville]{perez2018film}
Ethan Perez, Florian Strub, Harm De~Vries, Vincent Dumoulin, and Aaron
  Courville.
\newblock Film: Visual reasoning with a general conditioning layer.
\newblock In \emph{Thirty-Second AAAI Conference on Artificial Intelligence},
  2018.

\bibitem[Radford et~al.(2015)Radford, Metz, and
  Chintala]{radford2015unsupervised}
Alec Radford, Luke Metz, and Soumith Chintala.
\newblock Unsupervised representation learning with deep convolutional
  generative adversarial networks.
\newblock \emph{arXiv preprint arXiv:1511.06434}, 2015.

\bibitem[Romero \& Hoogendoorn(2019)Romero and Hoogendoorn]{romero2019co}
David~W Romero and Mark Hoogendoorn.
\newblock Co-attentive equivariant neural networks: Focusing equivariance on
  transformations co-occurring in data.
\newblock \emph{arXiv preprint arXiv:1911.07849}, 2019.

\bibitem[Romero et~al.(2020)Romero, Bekkers, Tomczak, and
  Hoogendoorn]{romero2020att}
David~W. Romero, Erik~J. Bekkers, Jakub~M. Tomczak, and Mark Hoogendoorn.
\newblock Attentive group equivariant convolutional networks, 2020.

\bibitem[Sabour et~al.(2017)Sabour, Frosst, and Hinton]{sabour2017dynamic}
Sara Sabour, Nicholas Frosst, and Geoffrey~E Hinton.
\newblock Dynamic routing between capsules.
\newblock In \emph{Advances in neural information processing systems}, pp.\
  3856--3866, 2017.

\bibitem[Schattschneider(1978)]{schattschneider1978plane}
Doris Schattschneider.
\newblock The plane symmetry groups: their recognition and notation.
\newblock \emph{The American Mathematical Monthly}, 85\penalty0 (6):\penalty0
  439--450, 1978.

\bibitem[Sifre \& Mallat(2013)Sifre and Mallat]{Sifre_2013_CVPR}
Laurent Sifre and Stephane Mallat.
\newblock Rotation, scaling and deformation invariant scattering for texture
  discrimination.
\newblock In \emph{Proceedings of the IEEE Conference on Computer Vision and
  Pattern Recognition (CVPR)}, June 2013.

\bibitem[Simonyan \& Zisserman(2014)Simonyan and Zisserman]{simonyan2014very}
Karen Simonyan and Andrew Zisserman.
\newblock Very deep convolutional networks for large-scale image recognition.
\newblock \emph{arXiv preprint arXiv:1409.1556}, 2014.

\bibitem[Sinha et~al.(2019)Sinha, Zhang, Goyal, Bengio, Larochelle, and
  Odena]{sinha2019small}
Samarth Sinha, Han Zhang, Anirudh Goyal, Yoshua Bengio, Hugo Larochelle, and
  Augustus Odena.
\newblock Small-gan: Speeding up gan training using core-sets.
\newblock \emph{arXiv preprint arXiv:1910.13540}, 2019.

\bibitem[Sinha et~al.(2020)Sinha, Goyal, Raffel, and Odena]{sinha2020top}
Samarth Sinha, Anirudh Goyal, Colin Raffel, and Augustus Odena.
\newblock Top-k training of gans: Improving generators by making critics less
  critical.
\newblock \emph{arXiv preprint arXiv:2002.06224}, 2020.

\bibitem[Szegedy et~al.(2016)Szegedy, Vanhoucke, Ioffe, Shlens, and
  Wojna]{szegedy2016rethinking}
Christian Szegedy, Vincent Vanhoucke, Sergey Ioffe, Jon Shlens, and Zbigniew
  Wojna.
\newblock Rethinking the inception architecture for computer vision.
\newblock In \emph{Proceedings of the IEEE conference on computer vision and
  pattern recognition}, pp.\  2818--2826, 2016.

\bibitem[Thanh-Tung \& Tran(2018)Thanh-Tung and Tran]{thanh2018catastrophic}
Hoang Thanh-Tung and Truyen Tran.
\newblock On catastrophic forgetting in generative adversarial networks.
\newblock \emph{arXiv preprint arXiv:1807.04015}, 2018.

\bibitem[Tran et~al.(2017)Tran, Ranganath, and Blei]{tran2017hierarchical}
Dustin Tran, Rajesh Ranganath, and David Blei.
\newblock Hierarchical implicit models and likelihood-free variational
  inference.
\newblock In \emph{Advances in Neural Information Processing Systems}, pp.\
  5523--5533, 2017.

\bibitem[Upadhyay \& Schrater(2018)Upadhyay and
  Schrater]{upadhyay2018generative}
Yash Upadhyay and Paul Schrater.
\newblock Generative adversarial network architectures for image synthesis
  using capsule networks.
\newblock \emph{arXiv preprint arXiv:1806.03796}, 2018.

\bibitem[Veeling et~al.(2018)Veeling, Linmans, Winkens, Cohen, and
  Welling]{veeling2018rotation}
Bastiaan~S Veeling, Jasper Linmans, Jim Winkens, Taco Cohen, and Max Welling.
\newblock Rotation equivariant cnns for digital pathology.
\newblock In \emph{International Conference on Medical image computing and
  computer-assisted intervention}, pp.\  210--218. Springer, 2018.

\bibitem[Wang et~al.(2018)Wang, Wu, Herranz, van~de Weijer, Gonzalez-Garcia,
  and Raducanu]{wang2018transferring}
Yaxing Wang, Chenshen Wu, Luis Herranz, Joost van~de Weijer, Abel
  Gonzalez-Garcia, and Bogdan Raducanu.
\newblock Transferring gans: generating images from limited data.
\newblock In \emph{Proceedings of the European Conference on Computer Vision
  (ECCV)}, pp.\  218--234, 2018.

\bibitem[Weiler \& Cesa(2019)Weiler and Cesa]{weiler2019general}
Maurice Weiler and Gabriele Cesa.
\newblock General e (2)-equivariant steerable cnns.
\newblock In \emph{Advances in Neural Information Processing Systems}, pp.\
  14334--14345, 2019.

\bibitem[White(2016)]{white2016sampling}
Tom White.
\newblock Sampling generative networks.
\newblock \emph{arXiv preprint arXiv:1609.04468}, 2016.

\bibitem[Winkels \& Cohen(2019)Winkels and Cohen]{winkels2019pulmonary}
Marysia Winkels and Taco~S Cohen.
\newblock Pulmonary nodule detection in ct scans with equivariant cnns.
\newblock \emph{Medical image analysis}, 55:\penalty0 15--26, 2019.

\bibitem[Wu et~al.(2019)Wu, Donahue, Balduzzi, Simonyan, and
  Lillicrap]{wu2019logan}
Yan Wu, Jeff Donahue, David Balduzzi, Karen Simonyan, and Timothy Lillicrap.
\newblock Logan: Latent optimisation for generative adversarial networks.
\newblock \emph{arXiv preprint arXiv:1912.00953}, 2019.

\bibitem[Zhang et~al.(2018)Zhang, Goodfellow, Metaxas, and
  Odena]{zhang2018self}
Han Zhang, Ian Goodfellow, Dimitris Metaxas, and Augustus Odena.
\newblock Self-attention generative adversarial networks.
\newblock \emph{arXiv preprint arXiv:1805.08318}, 2018.

\bibitem[Zhang et~al.(2020)Zhang, Zhang, Odena, and Lee]{Zhang2020Consistency}
Han Zhang, Zizhao Zhang, Augustus Odena, and Honglak Lee.
\newblock Consistency regularization for generative adversarial networks.
\newblock In \emph{International Conference on Learning Representations}, 2020.
\newblock URL \url{https://openreview.net/forum?id=S1lxKlSKPH}.

\bibitem[Zhao et~al.(2020{\natexlab{a}})Zhao, Cong, and
  Carin]{zhao2020leveraging}
Miaoyun Zhao, Yulai Cong, and Lawrence Carin.
\newblock On leveraging pretrained gans for limited-data generation.
\newblock \emph{arXiv preprint arXiv:2002.11810}, 2020{\natexlab{a}}.

\bibitem[Zhao et~al.(2020{\natexlab{b}})Zhao, Liu, Lin, Zhu, and
  Han]{zhao2020differentiable}
Shengyu Zhao, Zhijian Liu, Ji~Lin, Jun-Yan Zhu, and Song Han.
\newblock Differentiable augmentation for data-efficient gan training,
  2020{\natexlab{b}}.

\bibitem[Zhao et~al.(2020{\natexlab{c}})Zhao, Singh, Lee, Zhang, Odena, and
  Zhang]{zhao2020improved}
Zhengli Zhao, Sameer Singh, Honglak Lee, Zizhao Zhang, Augustus Odena, and Han
  Zhang.
\newblock Improved consistency regularization for gans, 2020{\natexlab{c}}.

\bibitem[Zhou et~al.(2019)Zhou, Liang, Song, Yu, Wang, Zhang, Yu, and
  Zhang]{pmlr-v97-zhou19c}
Zhiming Zhou, Jiadong Liang, Yuxuan Song, Lantao Yu, Hongwei Wang, Weinan
  Zhang, Yong Yu, and Zhihua Zhang.
\newblock {L}ipschitz generative adversarial nets.
\newblock In \emph{Proceedings of the 36th International Conference on Machine
  Learning}, pp.\  7584--7593, 2019.

\end{thebibliography}
\bibliographystyle{iclr2021_conference}

\newpage
\appendix

\section{Supplementary Results}
\label{app:suppl_res}

\begin{figure}[hp]
    \centering
    \includegraphics[width=1\textwidth]{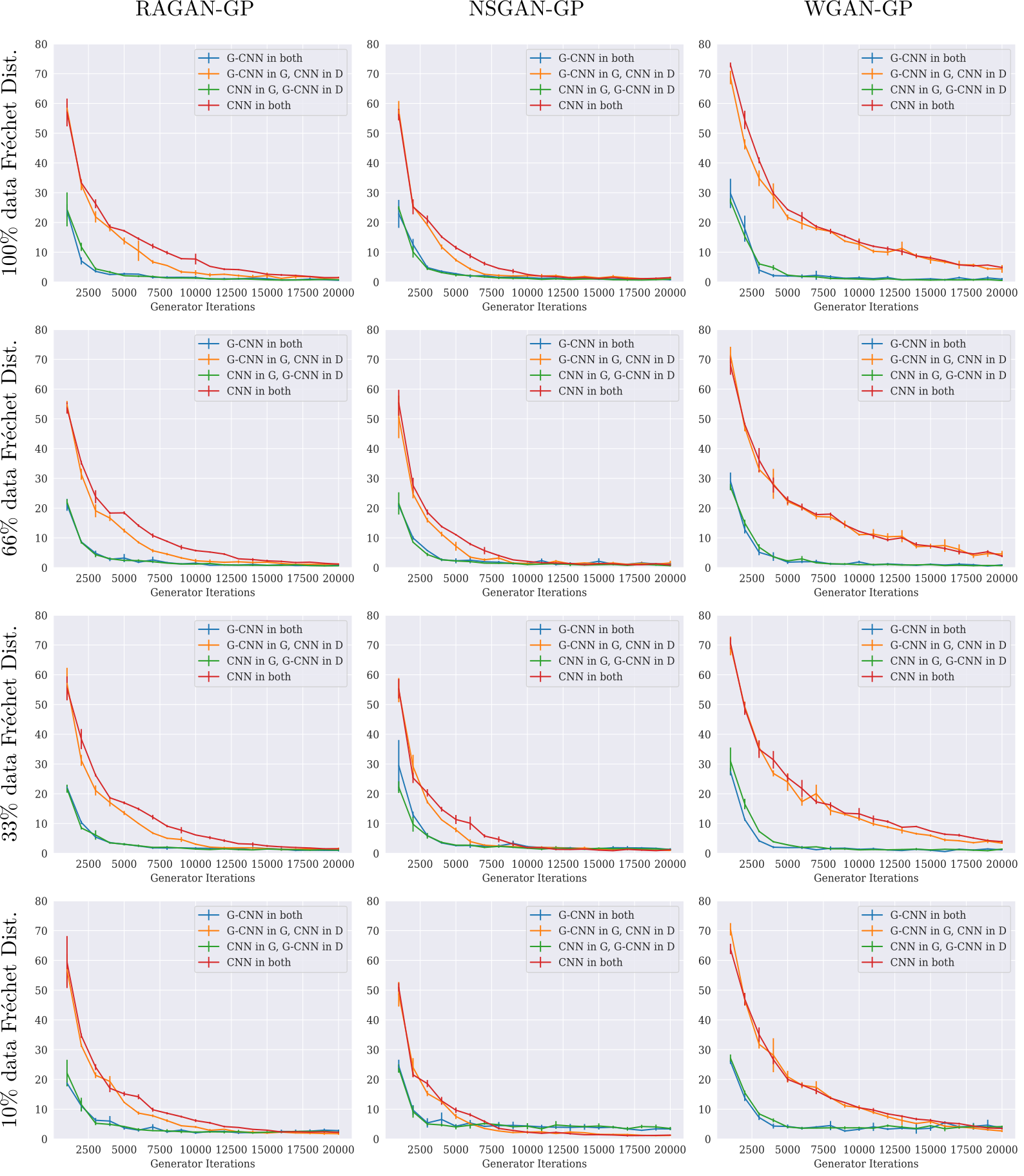}
    \caption{Convergence plots of all GAN ablation settings on Rotated MNIST across data availabilities (rows) and loss functions (columns). Fr\'echet distance to the validation set is evaluated every 1,000 generator iterations, for 20,000 iterations total. Experiments are repeated with 3 different random seeds and average trajectories are reported with standard deviation error bars. This figure is best interpreted alongside Table \ref{tab:rmnist} which lists best performances for each configuration.}
    \label{app:rmnist_res_fig}
\end{figure}

\begin{figure}[hp]
    \centering
    \includegraphics[height=0.9\textheight,width=\textwidth]{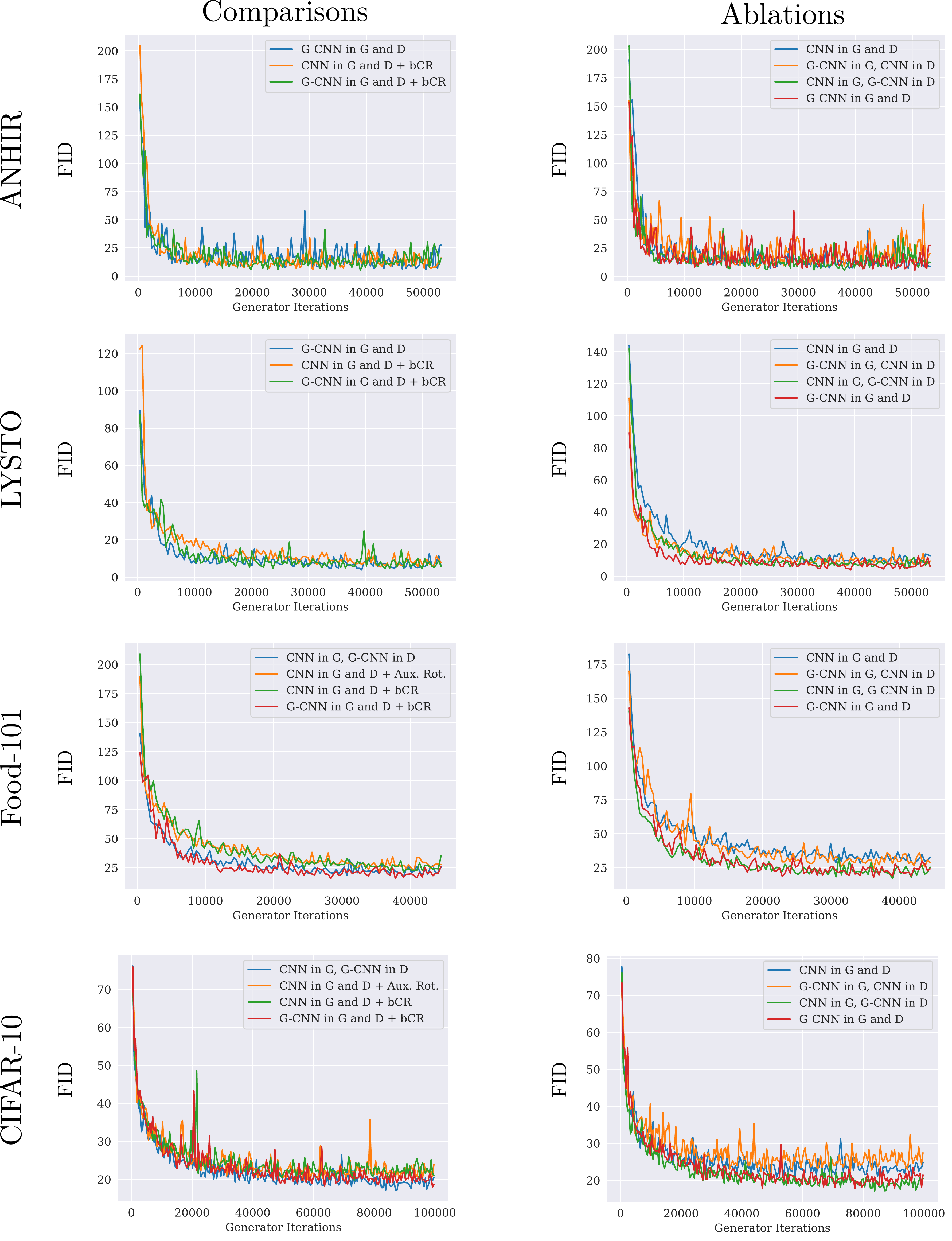}
    \caption{GAN convergence (FID vs. generator updates) for baseline comparisons of the best performing methods (\textbf{left}) and ablations (\textbf{right}) for all datasets. Readers are encouraged to zoom-in for better inspection.}
    \label{app:all_convergence_plots}
\end{figure}

\begin{figure}[!ph]
    \centering
    \includegraphics[width=1\textwidth]{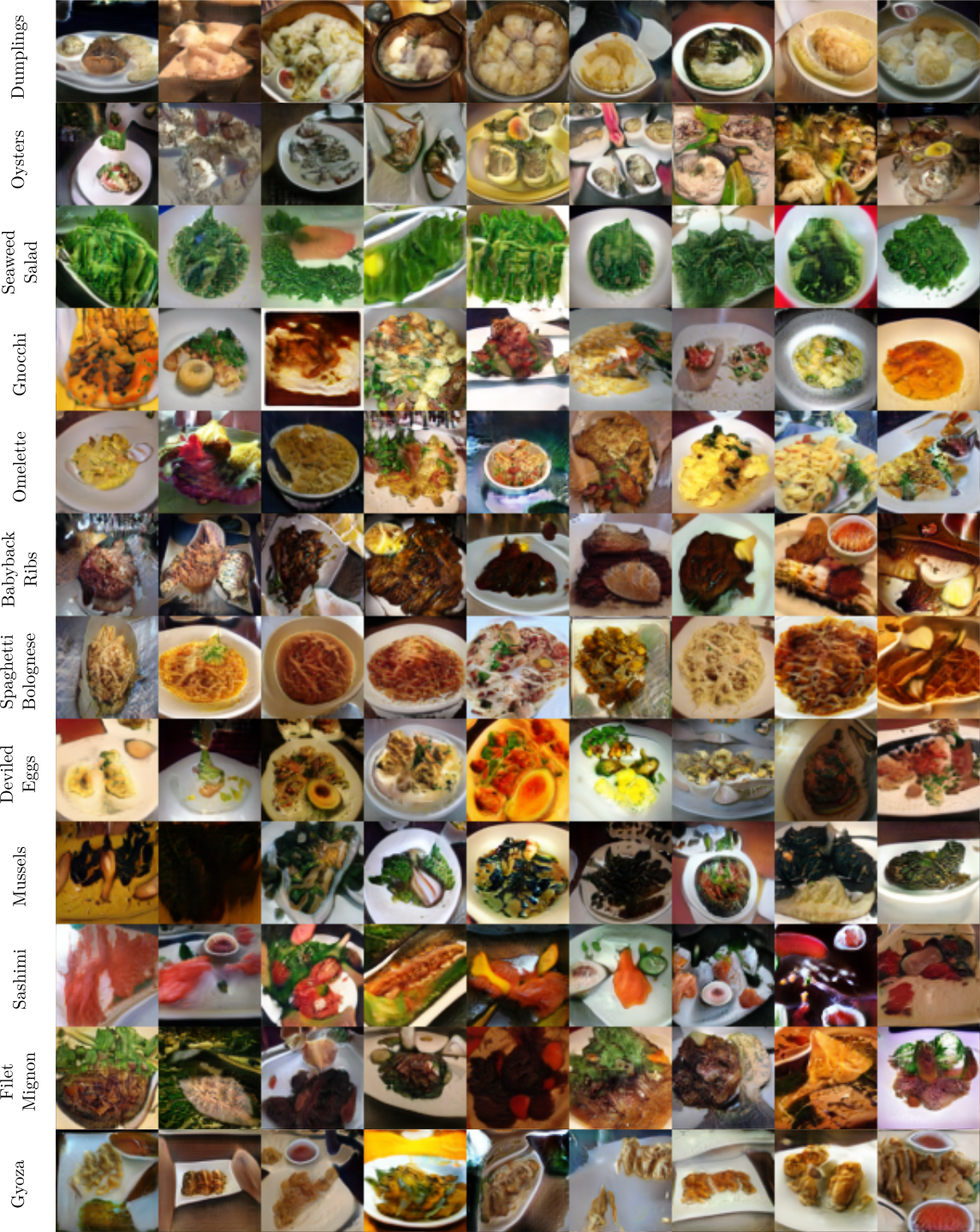}
    \caption{Random $64 \times 64$ Food-101 samples from arbitrarily chosen classes with no truncation taken from the best performing model snapshot with $p4$-equivariance (without augmentation) in the discriminator.}
    \label{app:food101_suppl_fig}
\end{figure}

\begin{figure}[hp]
    \centering
    \includegraphics[height=0.9\textheight]{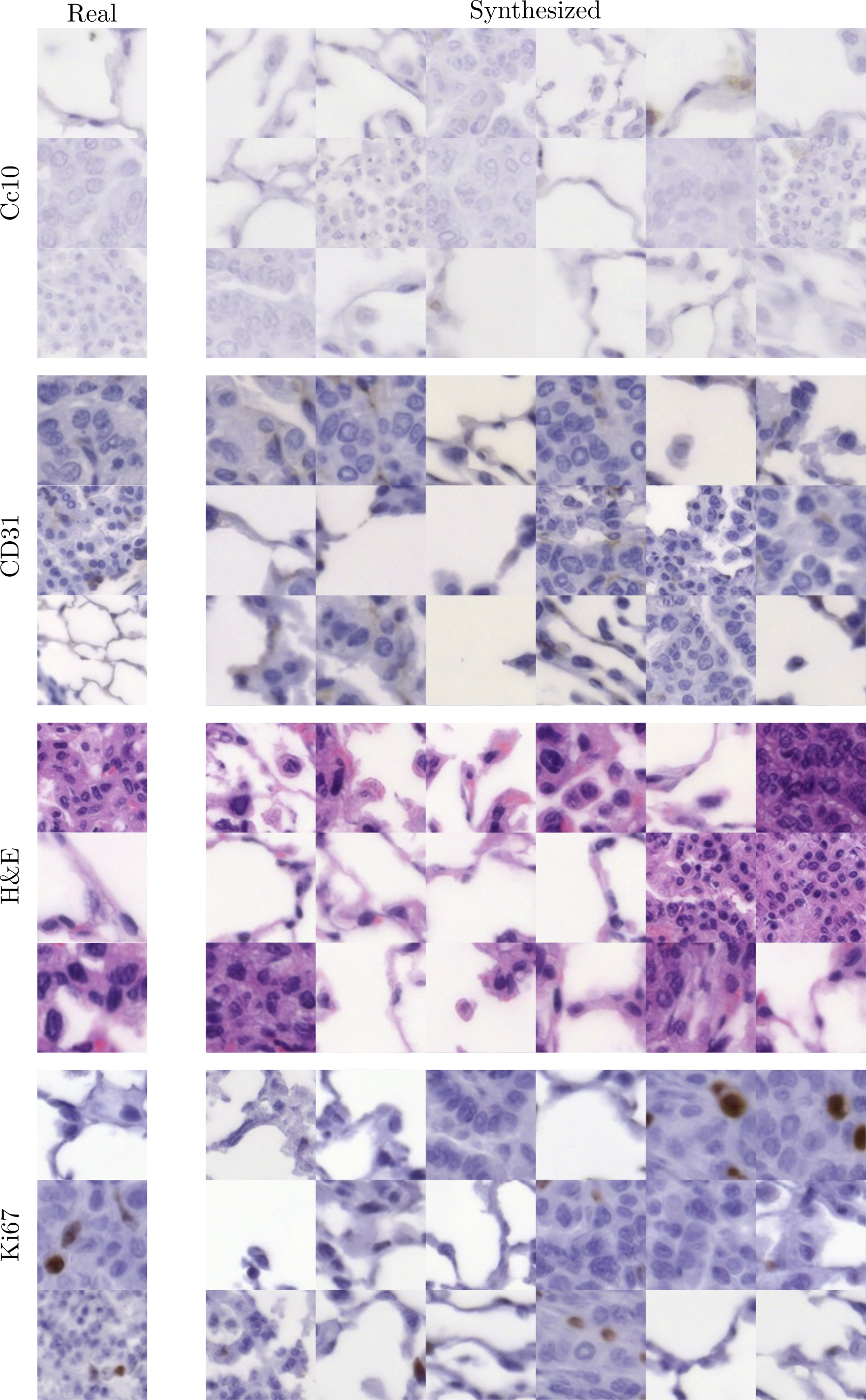}
    \caption{Random $128 \times 128$ ANHIR samples with no truncation taken from the best performing model snapshot with $p4m$-equivariance in both generator and discriminator (without augmentation). Selected real samples are shown in the left column for reference.}
    \label{app:anhir_suppl_fig}
\end{figure}

\begin{figure}[hp]
    \centering
    \includegraphics[height=0.9\textheight]{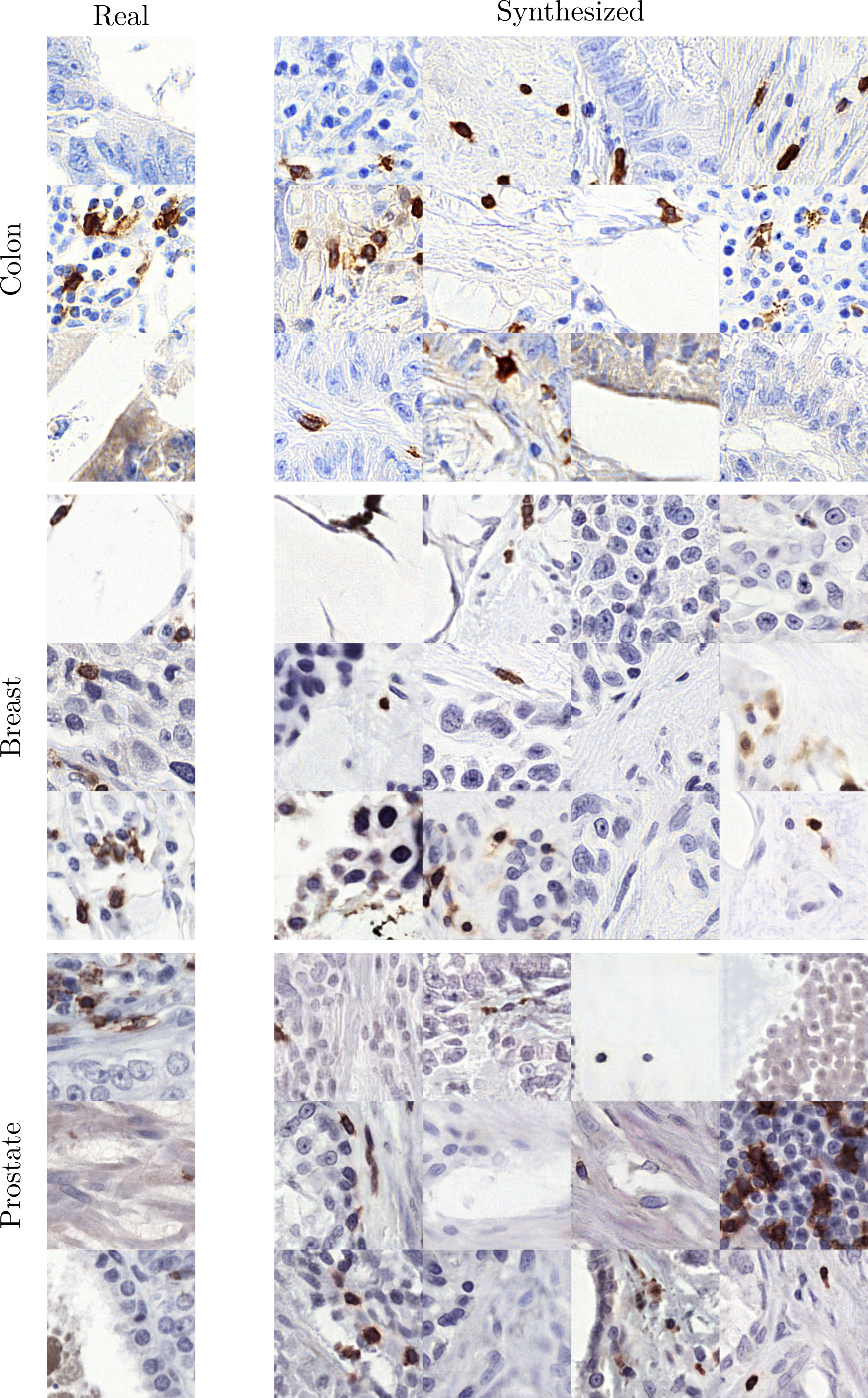}
    \caption{Random $256 \times 256$ LYSTO samples with no truncation taken from the best performing model snapshot with $p4m$-equivariance in both generator and discriminator (without augmentation). Selected real samples are shown in the left column for reference.}
    \label{app:lysto_suppl_fig}
\end{figure}

\begin{figure}[!ht]
    \centering
    \includegraphics[width=1\textwidth]{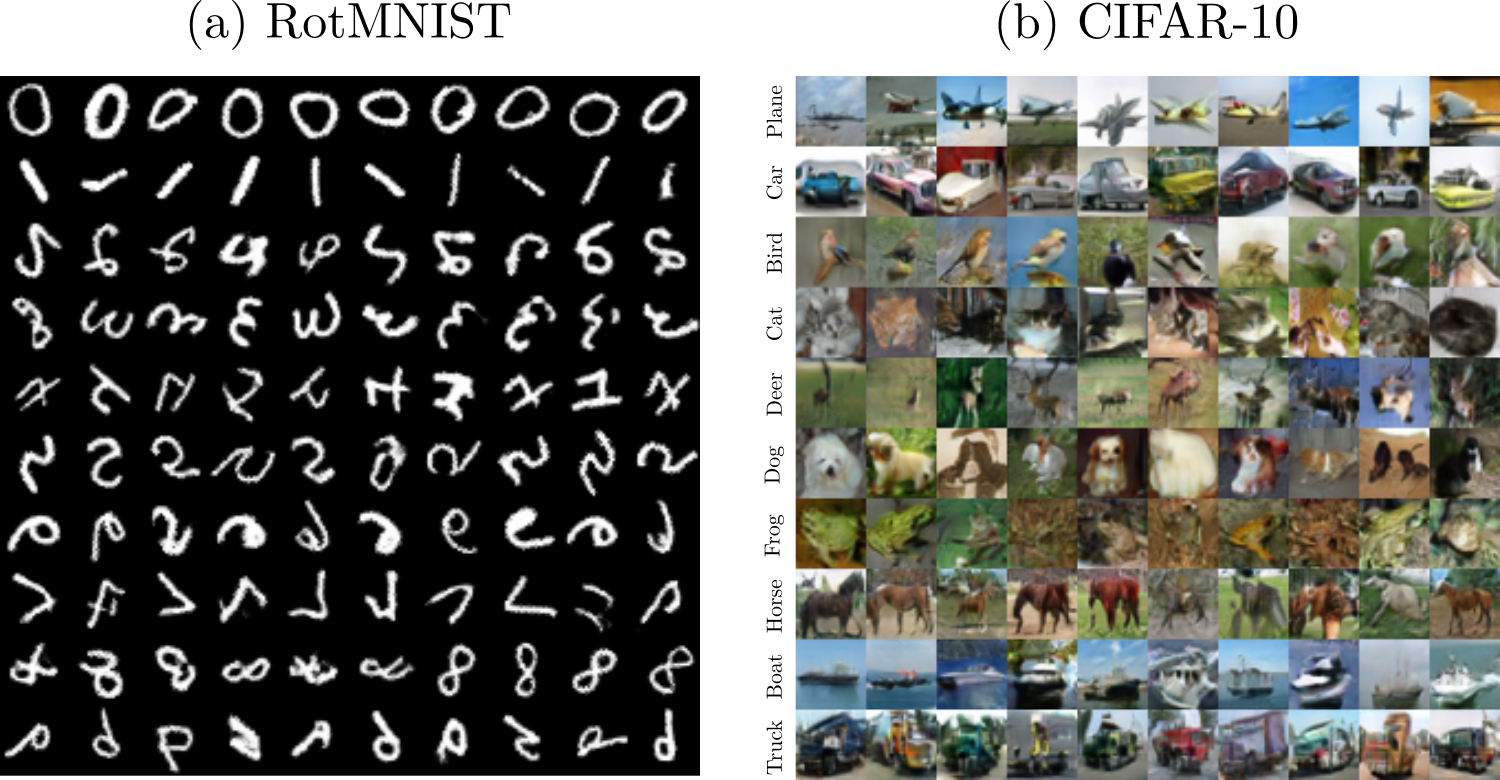}
    \caption{Random samples for RotMNIST ($28 \times 28$) and CIFAR-10 ($32 \times 32$) sampled with $\sigma = 0.75$ truncation trained without augmentation.}
    \label{app:cifar_rotmnist}
\end{figure}

\begin{table}[!h]
\begin{center}
\caption{Kernel Inception Distance results for Map2Sat translation on the Maps dataset. Lower is better.}
\begin{tabular}{lc}
 \toprule
 \textbf{Setting} & \textbf{KID} \\
 \midrule
 Pix2Pix \citep{isola2017image} & 0.1584 $\pm$ 0.0026 \\
 Pix2Pix \citep{isola2017image} (optimized) & 0.0663 $\pm$ 0.0038 \\
 CNN in G, G-CNN in D & 0.0333 $\pm$ 0.0005 \\
 G-CNN in G and D & 0.0399 $\pm$ 0.0024 \\
 \midrule
\end{tabular}
\label{app:maps_results}
\end{center}
\end{table}

\newpage
\section{Image-to-Image Translation}
To show the generic utility of equivariance in generative adversarial network tasks, we present a pilot study employing $p4$-equivariance in supervised image-to-image translation to learn mappings between visual domains. Using the popular \verb|Pix2Pix| model of \cite{isola2017image} as a baseline, we replace both networks with $p4$-equivariant models. For completeness, we also evaluate whether employing $p4$-equivariance in just the discriminator achieves comparable results to modifying both networks, as in the natural image datasets in the main text. 

We use the $256 \times 256$ \verb|Maps| dataset first introduced in \citep{isola2017image}, consisting of 1096 training and 1098 validation images of pairs of Google maps images and their corresponding satellite/aerial view images. As FID has a highly biased estimator, its use for evaluating generation with only 1098 validation samples is contraindicated \citep{BinkowskiSAG18}. We instead use the Kernel Inception Distance (KID) proposed by \cite{BinkowskiSAG18} which exhibits low bias for small sample sizes and is adopted in recent image translation studies \citep{kim2019u}. Briefly, as in FID, KID embeds real and fake images into the feature-space of an appropriately chosen network and computes the squared maximum-mean discrepancy (with a polynomial kernel) between their embeddings. Lower values of KID are better. We use the official \verb|Tensorflow| implementation and weights\footnote{\url{https://github.com/tensorflow/gan/blob/master/tensorflow_gan/python/eval/inception_metrics.py}}.

For baseline \verb|Pix2Pix|, we use pre-trained weights provided by the authors\footnote{\url{https://github.com/junyanz/pytorch-CycleGAN-and-pix2pix}}. Interestingly, we find that their architectures can be optimized for improved performance by replacing transposed convolutions with resize-convolutions, reducing the number of parameters by swapping $4 \times 4$ convolutional kernels for $3 \times 3$ kernels, and removing dropout. For equivariant models, we replace convolutions with $p4$-convolutions in this optimized architecture and halve the number of filters to keep the number of parameters similar across settings. Architectures are given in Tables \ref{arch:pix2pix_standard} and \ref{arch:pix2pix_GE}. We leave all other experimental details identical to \cite{isola2017image} for all models, such as training for 200 epochs with random crops under a cross-entropy GAN loss. 

\begin{figure}[t]
    \centering
    \includegraphics[width=0.95\textwidth]{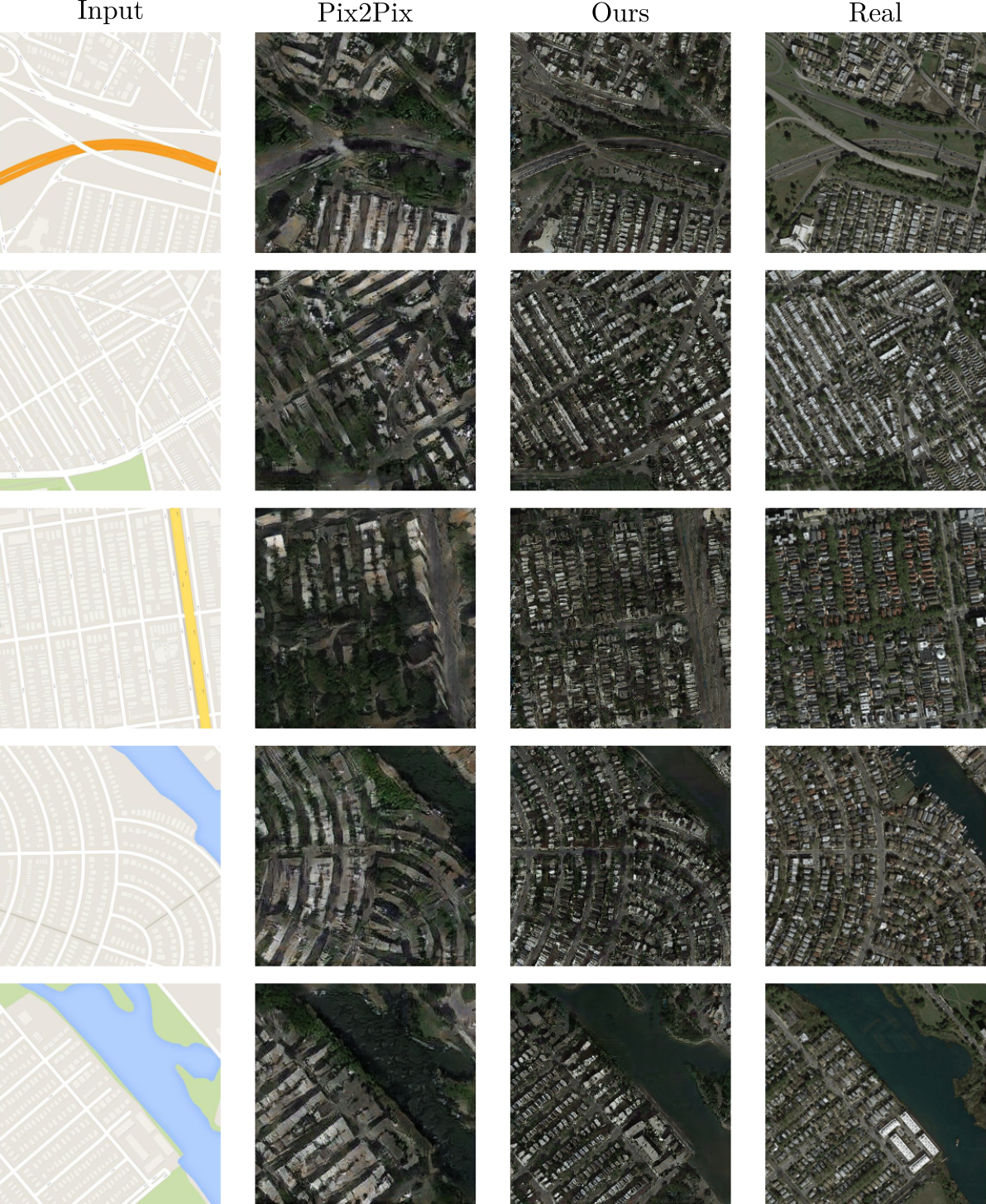}
    \caption{Arbitrarily selected sample translations from input map images (\textbf{Col. 1}) using either baseline Pix2Pix with publicly available pre-trained weights (\textbf{Col. 2}) or Pix2Pix with a $p4$-equivariant discriminator (\textbf{Col. 3}). Real aerial images are shown in Col. 4.}
    \label{app:pix2pix_results}
\end{figure}

Quantitative results are presented in Table \ref{app:maps_results} which shows that $p4$-equivariance in either setting improves over both original baseline and optimized baseline by a wide margin, with the best results coming from $p4$-equivariance in the discriminator alone. Qualitative results are presented in Figure \ref{app:pix2pix_results} showing improved translation fidelity, further supporting our hypothesis that equivariant networks benefit GAN tasks generically.

\section{Experimental details}
\label{app:exp_det}

\subsection{Data preparation}
\label{app:data_prep}
\subsubsection{LYSTO class conditioning}
\label{app:lysto_organ}

To validate the assumption of the organ source being a discriminative feature, a suitable test would be to train a classifier to distinguish between sources. We partition the original training set with a 60/40 train/test split. The original testing set is not used as it has no publicly available organ source information. The dataset has 3 classes - colon, breast, and prostate. Holding out 20\% of the new constructed training set for validation, we fine-tune ImageNet-pretrained VGG16 \citep{simonyan2014very} and achieve 98\% organ classification test accuracy, thus validating our assumption.

\subsubsection{ANHIR patch extraction}
\label{app:anhir_patch}

To extract patches for image synthesis, we choose the \verb|lung-lesion| images from the larger ANHIR dataset, as these images are provided at different scales and possess diverse staining. The images were cropped to the nearest multiples of 128, and $128 \times 128$ patches were then extracted. Foreground/background masking was performed via K-means clustering, followed by morphological dilation. The images were then gridded into $128 \times 128$ patches, i.e., there was no overlap between patches. If a patch contained less than 10\% foreground pixels, it was excluded from consideration.

\subsection{Additional Implementation Details} \label{app:addn_impl_details}

The following subsections list dataset-specific training strategies. Unless noted, all layers use orthogonal initializations. Batch normalization momentum is set to 0.1, and LeakyReLU slopes are set to 0.2 (if used). Spectral normalization is used everywhere except for the dense layer which learns the class embedding as specified in the BigGAN \verb|PyTorch| GitHub repository\footnote{\url{https://github.com/ajbrock/BigGAN-PyTorch}}.

For ablation studies, as GANs consist of two networks (the generator and discriminator), we replace group-equivariant layers (convolutional, normalization, and pooling) with the corresponding standard layers in either generator or discriminator to evaluate which network benefits the most from equivariant learning. When we remove equivariant layers from both networks, we recover our baseline comparison. All settings use roughly the same number of parameters, with a very small difference in parameter count arising from the $p4$ (or $p4m$) class-conditional batch normalization layers requiring fewer affine scale and shift parameters than their corresponding standard normalization layers. Tangentially, we note that the equivariant networks require higher amounts of computation time. For example, for a fixed number of training iterations on ANHIR, $p4m$-equivariant GANs currently require approximately four times the amount of computation time.

To identify a common shared stable hyperparameter configuration for all ablations of our method on real datasets, a grid search was performed for the ANHIR dataset over learning rates for generator and discriminator $(\eta_g, \eta_d) : (\{10^{-4}, 4\times 10^{-4}\}, \{5\times10^{-5}, 2\times 10^{-4}\})$, gradient penalty strengths ($\gamma = \{0.01, 0.1, 1.0, 10.0\}$), and binary choices as to whether to use batch normalization in the discriminator or not, whether to use average-pooling or max-pooling to reduce spatial extent in the discriminator, and whether to use a Gaussian latent space or a Bernoulli latent space. We use the identified hyperparameter configuration as an initial starting point for all datasets, modifying them as appropriate as described below.

For ANHIR, LYSTO, and Food-101 we use the relativistic average adversarial loss \citep{jolicoeur2018relativistic} for its stability and for CIFAR-10 we use the Hinge loss \citep{lim2017geometric,tran2017hierarchical} to remain consistent with the literature for that dataset. For our implementation of auxiliary rotations GAN \citep{chen2019self}, we use the suggested regularization weights. For balanced consistency regularization (bCR) \citep{zhao2020improved}, we find that dataset-specific tuning of the regularization strength was required.

\subsubsection{RotMNIST} \label{app:rmnist_impl}
Given the low resolution of Rotated MNIST, we take a straightforward approach to synthesis without residual connections. In the generator, we sample from a $64D$ Gaussian latent space, concatenate class embeddings, and linearly project as described in Section \ref{sec:gegans}. Four spectrally-normalized convolutional layers are then used with class-conditional batch normalization employed after every convolution except for the first and last layer. The discriminator uses three spectrally normalized convolutional layers, with leaky ReLU non-linearities. Average pooling is used to reduce the spatial extent of the feature maps, with global average pooling and conditional projection used at the end of the sequence. For NSGAN and RaGAN, we use the $R_1$ GP, conservatively setting $\gamma = 0.1$. For WGAN, we use the GP defined in \citet{gulrajani2017improved} to ensure the 1-Lipschitz constraint with the recommended weight of $10.0$. Learning rates were set to $\eta_G = 0.0001$ and $\eta_D = 0.0004$, respectively. For the $p4$-equivariant models, max-pooling over rotations is used after the last group-convolutional layer in both generator and discriminator to get planar feature maps. Architectures are presented in Tables \ref{arch:rmnist_standard} and \ref{arch:rmnist_GE}.

\subsubsection{ANHIR} \label{app:anhir_impl}
We sample from a $128D$ Gaussian latent space with a batch size of 32. The generator consists of 6 pre-activation residual blocks followed by a final convolutional layer to obtain a 3-channel output. We use class-conditional batch normalization after every convolution, except at the final layer. The discriminator uses 5 pre-activation residual blocks, followed by global average pooling and conditional projection. In the equivariant settings, we use residual blocks with $p4m$-convolutions for roto-reflective symmetries. We train with the relativisitic average loss and use the $R_1$ GP with $\gamma = 0.1$. Learning rates are set to $\eta_{G}=0.0001$ and $\eta_{D}=0.0004$. All models were trained for approximately 60,000 generator iterations. bCR weights for comparison were set to $\lambda_{real} = 0.1$ and $\lambda_{fake} = 0.05$ for roto-reflective augmentations, with higher values collapsing training. Architectures are presented in Tables \ref{arch:anhir_standard} and \ref{arch:anhir_GE}.

\subsubsection{LYSTO} \label{app:lysto_impl}
Implementation for LYSTO is similar to that of App. \ref{app:anhir_impl}, with some key differences due to the greater difficulty of training. Due to memory constraints, we use a batch size of 16. We increase the number of residual blocks to 6 in both generator and discriminator and halve the number of filters. The equivariant settings used the $p4m$ roto-reflective symmetries. We initially experienced low sample diversity across a variety of hyperparameter settings. Contrary to recent literature, we find that using batch normalization in the discriminator in addition to spectral normalization greatly improves training for this dataset. Further, halving the learning rates for both networks to $\eta_{G}=0.00005$ and $\eta_{D}=0.0002$ and increasing the strength of the gradient penalty to 1.0 were necessary for ensuring training stability.  As in App. \ref{app:anhir_impl}, all models were trained for approximately 60,000 generator iterations and bCR weights were set to $\lambda_{real} = 0.1$ and $\lambda_{fake} = 0.05$ for roto-reflective augmentations. As test set labels are not publicly available for LYSTO, we evaluate FID, Precision, and Recall to the training set itself as done in a subset of experiments within \citet{jolicoeur2018relativistic} and \citet{zhao2020differentiable}. Architectures are presented in Tables \ref{arch:lysto_standard} and \ref{arch:lysto_GE}.

\subsubsection{CIFAR-10} \label{app:cifar_impl}
For CIFAR-10, we make the following changes to our training parameters to be in accordance with prior art for BigGAN-like designs for this dataset: (1) layer weights are now initialized from $\mathcal{N}(0, 0.02)$; (2) average pooling is used in the discriminator instead of max pooling; (3) learning rates $\eta_G$ and $\eta_D$ are now equal and set to 0.0002; (4) the discriminator is updated four times per generator update; (5) architectures are modified as in Tables \ref{arch:cifar_standard} and \ref{arch:cifar_GE}; (6) we use the Hinge loss instead of the relativistic average loss. We use a batch size of 64. \citet{karras2020training} suggest an $R_1$ GP weight of $\gamma = 0.01$ for CIFAR-10 which we use here. We train all CIFAR-10 GANs for 100K generator iterations. bCR weights were set to $\lambda_{real} = 0.1$ and $\lambda_{fake} = 0.1$ for 90-degree rotation augmentations.

For the $p4$-equivariant discriminators, we move the pooling over the group to before the last residual block as stated in the main text. Alternatively, we experimented with using a single additional standard convolutional layer with 32 filters after the $p4$-residual blocks as a lightweight alternative to making an entire residual block non-equivariant but this worsened FID evaluation. Interestingly, we find that substituting Global Average Pooling for Global Sum Pooling in the CIFAR-10 discriminators lead to an improvement of $\sim$5 - 8 in terms of FID across the board. This architectural change to the ResNet-based GANs from \citet{gulrajani2017improved} was originally made in \citet{miyato2018spectral}, but to our knowledge has not been noted in the literature previously.

\subsubsection{Food-101}
Compared to the residual synthesis models in App. \ref{app:anhir_impl} and \ref{app:lysto_impl}, we make several changes. We sample from a $64D$ latent Gaussian to lower the number of dense parameters and substantially increase the width of the residual blocks to account for the high number of image classes. We find that an $8\times$ increase in the number of channels for the initial projection from the latent vector and class embedding improves training significantly. We use 4 residual blocks each in both generator and discriminator. For the equivariant setting, we use only $p4$ rotational symmetries to reduce training time. Importantly, we increase the batch size to 64 and the $R_1$ GP to $\gamma=1.0$, both of which improve the evaluation of all experimental settings. We train all GANs for $\sim$45K generator iterations. The suggested bCR weights of $\lambda_{real} = 10.0$ and $\lambda_{fake} = 10.0$ from \citet{zhao2020improved} were used here for 90-degree rotation augmentations. However, when bCR with default parameters was combined with $p4$-equivariance in \verb|G| and \verb|D|, augmentations start to `leak' into the generated images (e.g., \verb|G| generating upside-down plates), necessitating lower weights of $\lambda_{real} = 0.5$ and $\lambda_{fake} = 0.5$.

\subsection{Architectures}
\begin{figure}[t]
    \centering
    \includegraphics[width=1\textwidth]{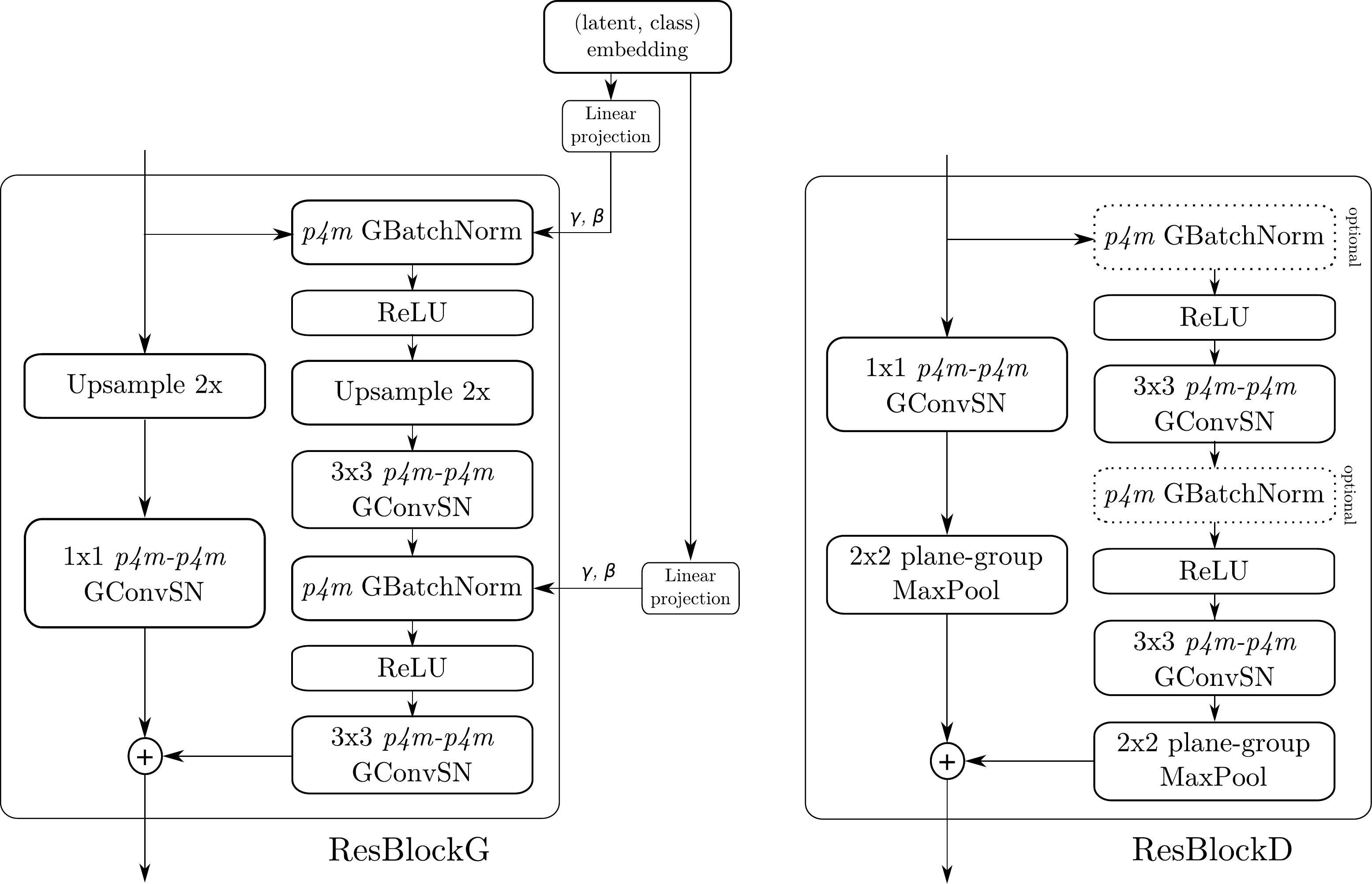}
    \caption{Residual blocks in the group-equivariant settings used in RGB image generation architectures. The choice of $p4$ or $p4m$ is dataset-specific. The generator uses ResBlockG (left) and the discriminator uses ResBlockD (right). The first residual block in the convolutional sequence in either network uses $z2$-$p4m$ group-convolutions for the initial layer. The non-equivariant settings replace all group-convolutions and normalizations within the residual blocks with standard techniques. Visual design inspired by \citet{brock2018large}.}
    \label{app:resblock_fig}
\end{figure}

Architectures for the Rotated MNIST experiments are given in Tables \ref{arch:rmnist_standard} and \ref{arch:rmnist_GE}, ANHIR in Tables \ref{arch:anhir_standard} and \ref{arch:anhir_GE}, and LYSTO in Tables \ref{arch:lysto_standard} and \ref{arch:lysto_GE}. The residual blocks used in the ANHIR, LYSTO, CIFAR-10, and Food-101 experiments are given in Figure \ref{app:resblock_fig}. \verb|SN| refers to spectral normalization and $(z2-p4), (p4-p4), (z2-p4m), (p4m-p4m)$ refer to the type of convolution used.

\newpage
\begin{table}
\parbox{.45\linewidth}{
\begin{center}
\begin{tabular}{c}
\toprule
Generator \\
\toprule
Sample $z \in \mathbb{R}^{64} \sim \mathcal{N}(0, I)$ \\
Embed $y \in \{0, ..., 9\}$ into $\hat y$ $\in \mathbb{R}^{64}$\\
Concatenate $z$ and $\hat y$ into $h \in \mathbb{R}^{128}$\\
\hline
Project and reshape $h$ to $7 \times 7 \times 128$ \\
\hline
$3 \times 3$ ConvSN, 128 $\rightarrow$ 512 \\
\hline
ReLU; Up $2\times$ \\
\hline
$3 \times 3$ ConvSN, 512 $\rightarrow$ 256 \\
\hline
CCBN$(\cdot,h)$; ReLU; Up $2\times$ \\
\hline
$3 \times 3$ ConvSN, 256 $\rightarrow$ 128\\
\hline
CCBN$(\cdot,h)$; ReLU \\
\hline
$3 \times 3$ ConvSN, 128 $\rightarrow$ 1 \\
\hline
tanh()\\
\bottomrule
\end{tabular}
\end{center}
}
\hfill
\parbox{.45\linewidth}{
\begin{center}
\begin{tabular}{c}
\toprule
Discriminator \\
\toprule
Input RGB image $x \in \mathbb{R}^{28\times 28 \times 1}$ \\
\hline
$3 \times 3$ ConvSN, $1 \rightarrow 128$\\
\hline
LeakyReLU, Avg. Pool \\
\hline
$3 \times 3$ ConvSN, $128 \rightarrow 256$\\
\hline
LeakyReLU, Avg. Pool \\
\hline
$3 \times 3$ ConvSN, $256 \rightarrow 512$\\
\hline
LeakyReLU, Avg. Pool \\
\hline
Global Average Pool into $f$ \\
\hline
Embed $y \in \{0, ..., 9\}$ into $\hat y'$\\
\hline
Projection step($\hat y', f$)  \\
\bottomrule
\end{tabular}
\end{center}
}
\caption{Architectures used for the standard generator and discriminator in the Rotated MNIST experiments.}
\label{arch:rmnist_standard}
\end{table}


\begin{table}
\parbox{.45\linewidth}{
\begin{center}
\begin{tabular}{c}
\toprule
Generator \\
\toprule
Sample $z \in \mathbb{R}^{64} \sim \mathcal{N}(0, I)$ \\
Embed $y \in \{0, ..., 9\}$ into $\hat y$ $\in \mathbb{R}^{64}$\\
Concatenate $z$ and $\hat y$ into $h \in \mathbb{R}^{128}$\\
\hline
Project and reshape $h$ to $7 \times 7 \times 128$ \\
\hline
$3 \times 3 \ z2-p4$ GConvSN, 128 $\rightarrow$ 256 \\
\hline
ReLU; Up $2\times$ \\
\hline
$3 \times 3 \ p4-p4$ GConvSN, 256 $\rightarrow$ 128 \\
\hline
CCBN$(\cdot,h)$; ReLU; Up $2\times$ \\
\hline
$3 \times 3 \ p4-p4$ GConvSN, 128 $\rightarrow$ 64\\
\hline
CCBN$(\cdot,h)$; ReLU \\
\hline
$3 \times 3 \ p4-p4$ GConvSN, 64 $\rightarrow$ 1 \\
\hline
$p4$-Max Pool \\
\hline
tanh()\\
\bottomrule
\end{tabular}
\end{center}
}
\hfill
\parbox{.45\linewidth}{
\begin{center}
\begin{tabular}{c}
\toprule
Discriminator \\
\toprule
Input RGB image $x \in \mathbb{R}^{28\times 28 \times 1}$ \\
\hline
$3 \times 3 \ z2-p4$ GConvSN, $1 \rightarrow 64$\\
\hline
LeakyReLU, Avg. Pool \\
\hline
$3 \times 3 \ p4-p4$ GConvSN, $64 \rightarrow 128$\\
\hline
LeakyReLU, Avg. Pool \\
\hline
$3 \times 3 \ p4-p4$ GConvSN, $128 \rightarrow 256$\\
\hline
LeakyReLU, Avg. Pool \\
\hline
$p4$-Max Pool \\
\hline
Global Average Pool into $f$ \\
\hline
Embed $y \in \{0, ..., 9\}$ into $\hat y'$\\
\hline
Projection step($\hat y', f$)  \\
\bottomrule
\end{tabular}
\end{center}
}
\caption{Architectures used for the $p4$-equivariant generator and discriminator in the Rotated MNIST experiments.}
\label{arch:rmnist_GE}
\end{table}


\begin{table}
\parbox{.45\linewidth}{
\begin{center}
\begin{tabular}{c}
\toprule
Generator \\
\toprule
Sample $z \in \mathbb{R}^{128} \sim \mathcal{N}(0, I)$ \\
Embed $y \in \{0, ..., 4\}$ into $\hat y$ $\in \mathbb{R}^{128}$\\
Concatenate $z$ and $\hat y$ into $h \in \mathbb{R}^{256}$\\
\hline
Project and reshape $h$ to $4 \times 4 \times 128$ \\
\hline
$z2-z2$ ResBlockG, 128 $\rightarrow$ 512 \\
\hline
$z2-z2$ ResBlockG, 512 $\rightarrow$ 256 \\
\hline
$z2-z2$ ResBlockG, 256 $\rightarrow$ 128 \\
\hline
$z2-z2$ ResBlockG, 128 $\rightarrow$ 64 \\
\hline
$z2-z2$ ResBlockG, 64 $\rightarrow$ 32 \\
\hline
BN; ReLU \\
\hline
$3 \times 3$ ConvSN, 32 $\rightarrow$ 3\\
\hline
tanh() \\
\bottomrule
\end{tabular}
\end{center}
}
\hfill
\parbox{.45\linewidth}{
\begin{center}
\begin{tabular}{c}
\toprule
Discriminator \\
\toprule
Input RGB image $x \in \mathbb{R}^{128\times 128 \times 3}$ \\
\hline
$z2-z2$ ResBlockD, 3 $\rightarrow$ 32 \\
\hline
$z2-z2$ ResBlockD, 32 $\rightarrow$ 64 \\
\hline
$z2-z2$ ResBlockD, 64 $\rightarrow$ 128 \\
\hline
$z2-z2$ ResBlockD, 128 $\rightarrow$ 256 \\
\hline
$z2-z2$ ResBlockD, 256 $\rightarrow$ 512 \\
\hline
ReLU\\
\hline
Global Average Pool into $f$ \\
\hline
Embed $y \in \{0, ..., 4\}$ into $\hat y'$\\
\hline
Projection step($\hat y', f$)  \\
\bottomrule
\end{tabular}
\end{center}
}
\caption{Architectures used for the standard generator and discriminator in the ANHIR experiments.}
\label{arch:anhir_standard}
\end{table}

\begin{table}
\parbox{.45\linewidth}{
\begin{center}
\begin{tabular}{c}
\toprule
Generator \\
\toprule
Sample $z \in \mathbb{R}^{128} \sim \mathcal{N}(0, I)$ \\
Embed $y \in \{0, ..., 4\}$ into $\hat y$ $\in \mathbb{R}^{128}$\\
Concatenate $z$ and $\hat y$ into $h \in \mathbb{R}^{256}$\\
\hline
Project and reshape $h$ to $4 \times 4 \times 128$ \\
\hline
$z2-p4m$ ResBlockG, 128 $\rightarrow$ 181 \\
\hline
$p4m-p4m$ ResBlockG, 181 $\rightarrow$ 90 \\
\hline
$p4m-p4m$ ResBlockG, 90 $\rightarrow$ 45 \\
\hline
$p4m-p4m$ ResBlockG, 45 $\rightarrow$ 22 \\
\hline
$p4m-p4m$ ResBlockG, 22 $\rightarrow$ 11 \\
\hline
$p4m$-BN; ReLU \\
\hline
$3 \times 3$ $p4m-p4m$ GConvSN, 11 $\rightarrow$ 3\\
\hline
$p4m$-Max Pool\\
\hline
tanh() \\
\bottomrule
\end{tabular}
\end{center}
}
\hfill
\parbox{.45\linewidth}{
\begin{center}
\begin{tabular}{c}
\toprule
Discriminator \\
\toprule
Input RGB image $x \in \mathbb{R}^{128\times 128 \times 3}$ \\
\hline
$z2-p4m$ ResBlockD, 3 $\rightarrow$ 11 \\
\hline
$p4m-p4m$ ResBlockD, 11 $\rightarrow$ 22 \\
\hline
$p4m-p4m$ ResBlockD, 22 $\rightarrow$ 45 \\
\hline
$p4m-p4m$ ResBlockD, 45 $\rightarrow$ 90 \\
\hline
$p4m-p4m$ ResBlockD, 90 $\rightarrow$ 181 \\
\hline
ReLU\\
\hline
$p4m$-Max Pool \\
\hline
Global Average Pool into $f$ \\
\hline
Embed $y \in \{0, ..., 4\}$ into $\hat y'$\\
\hline
Projection step($\hat y', f$)  \\
\bottomrule
\end{tabular}
\end{center}
}
\caption{Architectures used for the $p4m$-equivariant generator and discriminator in the ANHIR experiments.}
\label{arch:anhir_GE}
\end{table}

\begin{table}
\parbox{.45\linewidth}{
\begin{center}
\begin{tabular}{c}
\toprule
Generator \\
\toprule
Sample $z \in \mathbb{R}^{128} \sim \mathcal{N}(0, I)$ \\
Embed $y \in \{0, 1, 2\}$ into $\hat y$ $\in \mathbb{R}^{128}$\\
Concatenate $z$ and $\hat y$ into $h \in \mathbb{R}^{256}$\\
\hline
Project and reshape $h$ to $4 \times 4 \times 128$ \\
\hline
$z2-z2$ ResBlockG, 128 $\rightarrow$ 512 \\
\hline
$z2-z2$ ResBlockG, 512 $\rightarrow$ 256 \\
\hline
$z2-z2$ ResBlockG, 256 $\rightarrow$ 128 \\
\hline
$z2-z2$ ResBlockG, 128 $\rightarrow$ 64 \\
\hline
$z2-z2$ ResBlockG, 64 $\rightarrow$ 32 \\
\hline
$z2-z2$ ResBlockG, 32 $\rightarrow$ 16 \\
\hline
BN; ReLU \\
\hline
$3 \times 3$ ConvSN, 16 $\rightarrow$ 3\\
\hline
tanh() \\
\bottomrule
\end{tabular}
\end{center}
}
\hfill
\parbox{.45\linewidth}{
\begin{center}
\begin{tabular}{c}
\toprule
Discriminator \\
\toprule
Input RGB image $x \in \mathbb{R}^{256\times 256 \times 3}$ \\
\hline
$z2-z2$ ResBlockD-BN, 3 $\rightarrow$ 16 \\
\hline
$z2-z2$ ResBlockD-BN, 16 $\rightarrow$ 32 \\
\hline
$z2-z2$ ResBlockD-BN, 32 $\rightarrow$ 64 \\
\hline
$z2-z2$ ResBlockD-BN, 64 $\rightarrow$ 128 \\
\hline
$z2-z2$ ResBlockD-BN, 128 $\rightarrow$ 256 \\
\hline
$z2-z2$ ResBlockD-BN, 256 $\rightarrow$ 512 \\
\hline
ReLU\\
\hline
Global Average Pool into $f$ \\
\hline
Embed $y \in \{0, 1, 2\}$ into $\hat y'$\\
\hline
Projection step($\hat y', f$)  \\
\bottomrule
\end{tabular}
\end{center}
}
\caption{Architectures used for the standard generator and discriminator in the LYSTO experiments.}
\label{arch:lysto_standard}
\end{table}

\begin{table}
\parbox{.45\linewidth}{
\begin{center}
\begin{tabular}{c}
\toprule
Generator \\
\toprule
Sample $z \in \mathbb{R}^{128} \sim \mathcal{N}(0, I)$ \\
Embed $y \in \{0, 1, 2\}$ into $\hat y$ $\in \mathbb{R}^{128}$\\
Concatenate $z$ and $\hat y$ into $h \in \mathbb{R}^{256}$\\
\hline
Project and reshape $h$ to $4 \times 4 \times 128$ \\
\hline
$z2-p4m$ ResBlockG, 128 $\rightarrow$ 181 \\
\hline
$p4m-p4m$ ResBlockG, 181 $\rightarrow$ 90 \\
\hline
$p4m-p4m$ ResBlockG, 90 $\rightarrow$ 45 \\
\hline
$p4m-p4m$ ResBlockG, 45 $\rightarrow$ 22 \\
\hline
$p4m-p4m$ ResBlockG, 22 $\rightarrow$ 11 \\
\hline
$p4m-p4m$ ResBlockG, 11 $\rightarrow$ 5 \\
\hline
$p4m$-BN; ReLU \\
\hline
$3 \times 3$ $p4m-p4m$ GConvSN, 5 $\rightarrow$ 3\\
\hline
$p4m$-Max Pool\\
\hline
tanh() \\
\bottomrule
\end{tabular}
\end{center}
}
\hfill
\parbox{.45\linewidth}{
\begin{center}
\begin{tabular}{c}
\toprule
Discriminator \\
\toprule
Input RGB image $x \in \mathbb{R}^{256\times 256 \times 3}$ \\
\hline
$z2-p4m$ ResBlockD-BN, 3 $\rightarrow$ 5 \\
\hline
$p4m-p4m$ ResBlockD-BN, 5 $\rightarrow$ 11 \\
\hline
$p4m-p4m$ ResBlockD-BN, 11 $\rightarrow$ 22 \\
\hline
$p4m-p4m$ ResBlockD-BN, 22 $\rightarrow$ 45 \\
\hline
$p4m-p4m$ ResBlockD-BN, 45 $\rightarrow$ 90 \\
\hline
$p4m-p4m$ ResBlockD-BN, 90 $\rightarrow$ 181 \\
\hline
ReLU\\
\hline
$p4m$-Max Pool \\
\hline
Global Average Pool into $f$ \\
\hline
Embed $y \in \{0, 1, 2\}$ into $\hat y'$\\
\hline
Projection step($\hat y', f$)  \\
\bottomrule
\end{tabular}
\end{center}
}
\caption{Architectures used for the $p4m$-equivariant generator and discriminator in the LYSTO experiments.}
\label{arch:lysto_GE}
\end{table}

\begin{table}
\parbox{.45\linewidth}{
\begin{center}
\begin{tabular}{c}
\toprule
Generator \\
\toprule
Sample $z \in \mathbb{R}^{128} \sim \mathcal{N}(0, I)$ \\
Embed $y \in \{0, ..., 9\}$ into $\hat y$ $\in \mathbb{R}^{128}$\\
Concatenate $z$ and $\hat y$ into $h \in \mathbb{R}^{256}$\\
\hline
Project and reshape $h$ to $4 \times 4 \times 256$ \\
\hline
$z2-z2$ ResBlockG, 256 $\rightarrow$ 256 \\
\hline
$z2-z2$ ResBlockG, 256 $\rightarrow$ 256 \\
\hline
$z2-z2$ ResBlockG, 256 $\rightarrow$ 256 \\
\hline
BN; ReLU \\
\hline
$3 \times 3$ ConvSN, 256 $\rightarrow$ 3\\
\hline
tanh() \\
\bottomrule
\end{tabular}
\end{center}
}
\hfill
\parbox{.45\linewidth}{
\begin{center}
\begin{tabular}{c}
\toprule
Discriminator \\
\toprule
Input RGB image $x \in \mathbb{R}^{32\times 32 \times 3}$ \\
\hline
$z2-z2$ ResBlockD (avg. pool), \\
3 $\rightarrow$ 128 \\
\hline
$z2-z2$ ResBlockD (avg. pool), \\
128 $\rightarrow$ 128 \\
\hline
$z2-z2$ ResBlockD (no downsample), \\
128 $\rightarrow$ 128 \\
\hline
$z2-z2$ ResBlockD (no downsample), \\
128 $\rightarrow$ 128 \\
\hline
ReLU\\
\hline
Global Sum Pool into $f$ \\
\hline
Embed $y \in \{0, ..., 9\}$ into $\hat y'$\\
\hline
Projection step($\hat y', f$)  \\
\bottomrule
\end{tabular}
\end{center}
}
\caption{Architectures used for the standard generator and discriminator in the CIFAR-10 experiments.}
\label{arch:cifar_standard}
\end{table}

\begin{table}
\parbox{.45\linewidth}{
\begin{center}
\begin{tabular}{c}
\toprule
Generator \\
\toprule
Sample $z \in \mathbb{R}^{128} \sim \mathcal{N}(0, I)$ \\
Embed $y \in \{0, ..., 9\}$ into $\hat y$ $\in \mathbb{R}^{128}$\\
Concatenate $z$ and $\hat y$ into $h \in \mathbb{R}^{256}$\\
\hline
Project and reshape $h$ to $4 \times 4 \times 256$ \\
\hline
$z2-p4$ ResBlockG, 256 $\rightarrow$ 128 \\
\hline
$p4-p4$ ResBlockG, 128 $\rightarrow$ 128 \\
\hline
$p4-p4$ ResBlockG, 128 $\rightarrow$ 128 \\
\hline
$p4$-BN; ReLU \\
\hline
$3 \times 3$ $p4-p4$ GConvSN, 128 $\rightarrow$ 3\\
\hline
$p4$-Max Pool\\
\hline
tanh() \\
\bottomrule
\end{tabular}
\end{center}
}
\hfill
\parbox{.45\linewidth}{
\begin{center}
\begin{tabular}{c}
\toprule
Discriminator \\
\toprule
Input RGB image $x \in \mathbb{R}^{64\times 64 \times 3}$ \\
\hline
$z2-p4$ ResBlockD (avg. pool), \\
3 $\rightarrow$ 64 \\
\hline
$z2-p4$ ResBlockD (avg. pool), \\
64 $\rightarrow$ 64 \\
\hline
$p4-p4$ ResBlockD (no downsample), \\
64 $\rightarrow$ 64 \\
\hline
$p4$-Max Pool \\
\hline
$z2-z2$ ResBlockD (no downsample), \\
64 $\rightarrow$ 128 \\
\hline
ReLU\\
\hline
Global Sum Pool into $f$ \\
\hline
Embed $y \in \{0, ..., 9\}$ into $\hat y'$\\
\hline
Projection step($\hat y', f$)  \\
\bottomrule
\end{tabular}
\end{center}
}
\caption{Architectures used for the $p4$-equivariant generator and discriminator in the CIFAR-10 experiments.}
\label{arch:cifar_GE}
\end{table}

\begin{table}
\parbox{.45\linewidth}{
\begin{center}
\begin{tabular}{c}
\toprule
Generator \\
\toprule
Sample $z \in \mathbb{R}^{64} \sim \mathcal{N}(0, I)$ \\
Embed $y \in \{0, ..., 100\}$ into $\hat y$ $\in \mathbb{R}^{64}$\\
Concatenate $z$ and $\hat y$ into $h \in \mathbb{R}^{128}$\\
\hline
Project and reshape $h$ to $4 \times 4 \times 1024$ \\
\hline
$z2-z2$ ResBlockG, 1024 $\rightarrow$ 512 \\
\hline
$z2-z2$ ResBlockG, 512 $\rightarrow$ 384 \\
\hline
$z2-z2$ ResBlockG, 384 $\rightarrow$ 256 \\
\hline
$z2-z2$ ResBlockG, 256 $\rightarrow$ 192 \\
\hline
BN; ReLU \\
\hline
$3 \times 3$ ConvSN, 192 $\rightarrow$ 3\\
\hline
tanh() \\
\bottomrule
\end{tabular}
\end{center}
}
\hfill
\parbox{.45\linewidth}{
\begin{center}
\begin{tabular}{c}
\toprule
Discriminator \\
\toprule
Input RGB image $x \in \mathbb{R}^{64\times 64 \times 3}$ \\
\hline
$z2-z2$ ResBlockD, 3 $\rightarrow$ 128 \\
\hline
$z2-z2$ ResBlockD, 128 $\rightarrow$ 256 \\
\hline
$z2-z2$ ResBlockD, 256 $\rightarrow$ 512 \\
\hline
$z2-z2$ ResBlockD, 512 $\rightarrow$ 784 \\
\hline
ReLU\\
\hline
Global Average Pool into $f$ \\
\hline
Embed $y \in \{0, ..., 100\}$ into $\hat y'$\\
\hline
Projection step($\hat y', f$)  \\
\bottomrule
\end{tabular}
\end{center}
}
\caption{Architectures used for the standard generator and discriminator in the Food-101 experiments.}
\label{arch:food_standard}
\end{table}

\begin{table}
\parbox{.45\linewidth}{
\begin{center}
\begin{tabular}{c}
\toprule
Generator \\
\toprule
Sample $z \in \mathbb{R}^{64} \sim \mathcal{N}(0, I)$ \\
Embed $y \in \{0, ..., 100\}$ into $\hat y$ $\in \mathbb{R}^{64}$\\
Concatenate $z$ and $\hat y$ into $h \in \mathbb{R}^{128}$\\
\hline
Project and reshape $h$ to $4 \times 4 \times 1024$ \\
\hline
$z2-p4$ ResBlockG, 1024 $\rightarrow$ 256 \\
\hline
$p4-p4$ ResBlockG, 256 $\rightarrow$ 192 \\
\hline
$p4-p4$ ResBlockG, 192 $\rightarrow$ 128 \\
\hline
$p4-p4$ ResBlockG, 128 $\rightarrow$ 96 \\
\hline
$p4$-BN; ReLU \\
\hline
$3 \times 3$ $p4-p4$ GConvSN, 96 $\rightarrow$ 3\\
\hline
$p4$-Max Pool\\
\hline
tanh() \\
\bottomrule
\end{tabular}
\end{center}
}
\hfill
\parbox{.45\linewidth}{
\begin{center}
\begin{tabular}{c}
\toprule
Discriminator \\
\toprule
Input RGB image $x \in \mathbb{R}^{64\times 64 \times 3}$ \\
\hline
$z2-p4$ ResBlockD, 3 $\rightarrow$ 64 \\
\hline
$p4-p4$ ResBlockD, 64 $\rightarrow$ 128 \\
\hline
$p4-p4$ ResBlockD, 128 $\rightarrow$ 256 \\
\hline
$p4$-Max Pool \\
\hline
$z2-z2$ ResBlockD, 256 $\rightarrow$ 784 \\
\hline
ReLU\\
\hline
Global Average Pool into $f$ \\
\hline
Embed $y \in \{0, ..., 100\}$ into $\hat y'$\\
\hline
Projection step($\hat y', f$)  \\
\bottomrule
\end{tabular}
\end{center}
}
\caption{Architectures used for the $p4$-equivariant generator and discriminator in the Food-101 experiments.}
\label{arch:food_GE}
\end{table}

\begin{table}
\parbox{.45\linewidth}{
\begin{center}
\begin{tabular}{c}
\toprule
Generator \\
\toprule
Input RGB image $x \in \mathbb{R}^{256\times 256 \times 3}$\\
\hline
$h_1:$ $z2-z2$ DownBlock, 3 $\rightarrow$ 64\\
\hline
$h_2:$ $z2-z2$ DownBlock, 64 $\rightarrow$ 128\\
\hline
$h_3:$ $z2-z2$ DownBlock, 128 $\rightarrow$ 256\\
\hline
$h_4:$ $z2-z2$ DownBlock, 512 $\rightarrow$ 512\\
\hline
$h_5:$ $z2-z2$ DownBlock, 512 $\rightarrow$ 512\\
\hline
$h_6:$ $z2-z2$ DownBlock, 512 $\rightarrow$ 512\\
\hline
$h_7:$ $z2-z2$ DownBlock, 512 $\rightarrow$ 512\\
\hline
$h_8:$ $z2-z2$ DownBlock, 512 $\rightarrow$ 512\\
\hline
$z2-z2$ UpBlock, 512 $\rightarrow$ 512; Concatenate $h_7$\\
\hline
$z2-z2$ UpBlock, 512 $\rightarrow$ 512; Concatenate $h_6$\\
\hline
$z2-z2$ UpBlock, 512 $\rightarrow$ 512; Concatenate $h_5$\\
\hline
$z2-z2$ UpBlock, 512 $\rightarrow$ 512; Concatenate $h_4$\\
\hline
$z2-z2$ UpBlock, 512 $\rightarrow$ 256; Concatenate $h_3$\\
\hline
$z2-z2$ UpBlock, 256 $\rightarrow$ 128; Concatenate $h_2$\\
\hline
$z2-z2$ UpBlock, 128 $\rightarrow$ 64; Concatenate $h_1$\\
\hline
Upsample $2\times$, $3 \times 3$ Conv, 64 $\rightarrow$ 3\\
\hline
tanh() \\
\bottomrule
\end{tabular}
\end{center}
}
\hfill
\parbox{.45\linewidth}{
\begin{center}
\begin{tabular}{c}
\toprule
Discriminator \\
\toprule
Input RGB image $x \in \mathbb{R}^{256\times 256 \times 3}$ \\
Input RGB image $y \in \mathbb{R}^{256\times 256 \times 3}$ \\
Concatenate $x$ and $y$ feature-wise \\
\hline
$z2-z2$ DownBlock, 3 $\rightarrow$ 64\\
\hline
$z2-z2$ DownBlock, 64 $\rightarrow$ 128\\
\hline
$z2-z2$ DownBlock, 128 $\rightarrow$ 256\\
\hline
$z2-z2$ $3\times3$ Conv 256 $\rightarrow$ 512,\\
BatchNorm, Leaky ReLU \\
\hline
$z2-z2$ $3\times3$ Conv 512 $\rightarrow$ 1,\\
\hline
tanh() \\
\bottomrule
\end{tabular}
\end{center}
}
\caption{Architectures used for the standard generator and discriminator in the Pix2Pix experiments. Each DownBlock consists of a $3\times3$ Convolution, $2\times$ Average Pool, Batch Normalization, and Leaky ReLU activation. Each UpBlock consists of $2\times$ nearest-neighbors upsampling, $3\times3$ Convolution, Batch Normalization, and Leaky ReLU activation.}
\label{arch:pix2pix_standard}
\end{table}

\begin{table}
\parbox{.45\linewidth}{
\begin{center}
\begin{tabular}{c}
\toprule
Generator \\
\toprule
Input RGB image $x \in \mathbb{R}^{256\times 256 \times 3}$\\
\hline
$h_1:$ $z2-p4$ DownBlock, 3 $\rightarrow$ 64\\
\hline
$h_2:$ $p4-p4$ DownBlock, 64 $\rightarrow$ 128\\
\hline
$h_3:$ $p4-p4$ DownBlock, 128 $\rightarrow$ 256\\
\hline
$h_4:$ $p4-p4$ DownBlock, 512 $\rightarrow$ 512\\
\hline
$h_5:$ $p4-p4$ DownBlock, 512 $\rightarrow$ 512\\
\hline
$h_6:$ $p4-p4$ DownBlock, 512 $\rightarrow$ 512\\
\hline
$h_7:$ $p4-p4$ DownBlock, 512 $\rightarrow$ 512\\
\hline
$h_8:$ $p4-p4$ DownBlock, 512 $\rightarrow$ 512\\
\hline
$p4-p4$ UpBlock, 512 $\rightarrow$ 512; Concatenate $h_7$\\
\hline
$p4-p4$ UpBlock, 512 $\rightarrow$ 512; Concatenate $h_6$\\
\hline
$p4-p4$ UpBlock, 512 $\rightarrow$ 512; Concatenate $h_5$\\
\hline
$p4-p4$ UpBlock, 512 $\rightarrow$ 512; Concatenate $h_4$\\
\hline
$p4-p4$ UpBlock, 512 $\rightarrow$ 256; Concatenate $h_3$\\
\hline
$p4-p4$ UpBlock, 256 $\rightarrow$ 128; Concatenate $h_2$\\
\hline
$p4-p4$ UpBlock, 128 $\rightarrow$ 64; Concatenate $h_1$\\
\hline
Upsample $2\times$, $3 \times 3$ GConv, 64 $\rightarrow$ 3\\
\hline
$p4$-average pool, tanh() \\
\bottomrule
\end{tabular}
\end{center}
}
\hfill
\parbox{.45\linewidth}{
\begin{center}
\begin{tabular}{c}
\toprule
Discriminator \\
\toprule
Input RGB image $x \in \mathbb{R}^{256\times 256 \times 3}$ \\
Input RGB image $y \in \mathbb{R}^{256\times 256 \times 3}$ \\
Concatenate $x$ and $y$ feature-wise \\
\hline
$z2-p4$ DownBlock, 3 $\rightarrow$ 64\\
\hline
$p4-p4$ DownBlock, 64 $\rightarrow$ 128\\
\hline
$p4-p4$ DownBlock, 128 $\rightarrow$ 256\\
\hline
$p4-p4$ $3\times3$ Conv 256 $\rightarrow$ 512,\\
BatchNorm, Leaky ReLU \\
\hline
$p4-p4$ $3\times3$ Conv 512 $\rightarrow$ 1,\\
\hline
$p4$-average pool \\
\hline 
tanh() \\
\bottomrule
\end{tabular}
\end{center}
}
\caption{Architectures used for the $p4$-equivariant generator and discriminator in the Pix2Pix experiments. Each DownBlock consists of a $3\times3$ $p4$-convolution, $2\times$ Average Pool, $p4$-Batch Normalization, and Leaky ReLU activation. Each UpBlock consists of $2\times$ nearest-neighbors upsampling, $3\times3$ $p4$-Convolution, $p4$-Batch Normalization, and Leaky ReLU activation.}
\label{arch:pix2pix_GE}
\end{table}

\end{document}